\documentclass[10pt,twocolumn,letterpaper]{article}

\usepackage[pagenumbers]{cvpr} % To force page numbers, e.g. for an arXiv version

\usepackage{listings}
\usepackage[table]{xcolor}
\usepackage{multirow}
\usepackage{multicol}
\usepackage{makecell}
\usepackage{graphicx}
\usepackage{amsmath}
\usepackage{amssymb}
\usepackage{booktabs}
\usepackage[marginal]{footmisc}

\usepackage{marvosym}

\usepackage[pagebackref,breaklinks,colorlinks]{hyperref}
\usepackage[capitalize]{cleveref}
\crefname{section}{Sec.}{Secs.}
\Crefname{section}{Section}{Sections}
\Crefname{table}{Table}{Tables}
\crefname{table}{Tab.}{Tabs.}

\def\cvprPaperID{**}
\def\confName{****}
\def\confYear{****}

\renewcommand{\paragraph}[1]{\vspace{2.0mm}\noindent\textbf{#1}}

\newcommand{\authorskip}{\hspace{2.5mm}}

\begin{document}
%%%%%%%%% TITLE - PLEASE UPDATE
\title{Position Embedding Needs an Independent Layer Normalization}
%Improving Vision Transformers with Layer-adaptive Position Embedding

% \author{Runyi Yu$^{1*}$ \authorskip Zhennan Wang$^{2*}$ \authorskip Yinhuai Wang$^{1*}$ \authorskip Kehan Li$^{1}$ \authorskip Yian Zhao$^{1}$ \\
% Jian Zhang$^{1,2}$ \authorskip Guoli Song$^{2}$\textsuperscript{\Letter} \authorskip Jie Chen$^{1,2}$\textsuperscript{\Letter}\\ [2mm]

\author{Runyi Yu$^{1}$ \authorskip Zhennan Wang$^{2}$ \authorskip Yinhuai Wang$^{1}$ \authorskip Kehan Li$^{1}$ \authorskip Yian Zhao$^{1}$ \\
Jian Zhang$^{1,2}$ \authorskip Guoli Song$^{2}$ \authorskip Jie Chen$^{1,2}$\\ [2mm]
$^{1}$ School of Electronic and Computer Engineering, Peking University, China\\ [2mm]
$^{2}$ Peng Cheng Laboratory, Shenzhen, China\\}
\maketitle

% \footnote{$^{*}$ Equal contribution.}
% \footnote{\textsuperscript{\Letter} Corresponding author: Guoli Song, Jie Chen.}

%%%%%%%%% ABSTRACT
% \begin{abstract}
%   Vision Transformer (VT) is one of the most popular research focueses. In existing VTs, the most common way to add position information to is adding Postion Embedding (PE) and Patch Embedding together before entering the encoder. However, this kind of PE cannot provide VTs with adaptable position information according to the different requirements of each layer, which results in limited performance. Specifically, the shallow layers need more local position information, and the deep layers need more global position information. To overcome this limitation, we propose an adaptive PE method, {\bf LaPE}, which is implemented by adding the layer normalized PE ahead of the Muti-Head Self-Attention module of each layer. The per-channel affine transformation coefficients of LN can adjust the position information expressed by PE, and the coefficients are different among layers, which can provide hierarchical position information. Our method is very simple and effect, and can serve as a basic performance enhancement method for VTs. Extensive experiments demonstrate that our method can improve all kinds of VTs on different datasets. For example, LaPE improves 0.94\% for ViT\_Lite on Cifar10, 0.98\% for CCT on Cifar100, and 1.30\% for DeiT on ImageNet.
% \end{abstract}

\begin{abstract}
The Position Embedding (PE) is critical for Vision Transformers (VTs) due to the permutation-invariance of self-attention operation. By analyzing the input and output of each encoder layer in VTs using reparameterization and visualization, we find that the default PE joining method (simply adding the PE and patch embedding together) operates the same affine transformation to token embedding and PE, which limits the expressiveness of PE and hence constrains the performance of VTs.
To overcome this limitation, we propose a simple, effective, and robust method. Specifically, we provide two independent layer normalizations for token embeddings and PE for each layer, and add them together as the input of each layer’s Muti-Head Self-Attention module. Since the method allows the model to adaptively adjust the information of PE for different layers, we name it as \textbf{L}ayer-\textbf{a}daptive \textbf{P}osition \textbf{E}mbedding, abbreviated as LaPE. Extensive experiments demonstrate that LaPE can improve various VTs with different types of PE and make VTs robust to PE types. For example, LaPE improves 0.94\% accuracy for ViT-Lite on Cifar10, 0.98\% for CCT on Cifar100, and 1.72\% for DeiT on ImageNet-1K, which is remarkable considering the negligible extra parameters, memory and computational cost brought by LaPE. The code is publicly available at \url{https://github.com/Ingrid725/LaPE}.
\end{abstract}

%------------------------------------------------------------------------
%------------------------------------------------------------------------

%%%%%%%%% BODY TEXT
\section{Introduction}
\label{sec:intro}
Recently, Vision Transformer (VT) has become one of the most popular research topics due to its superior performance on various computer vision tasks, such as image classification, detection, and segmentation. ViT\cite{dosovitskiy2020image} is the first pure transformer model for image classification, which outperforms CNNs when applied to large training data. Since then, many works based on ViT\cite{dosovitskiy2020image} have sprung up. Lots of work improves the tokenization\cite{hassani2021escaping,yuan2021tokens}, self-attention mechanism\cite{liu2021swin,zhang2022nested,yuan2022volo,wang2022uformer}, architecuture\cite{wang2021pyramid,touvron2021training,yuan2021incorporating,wu2021cvt,Mao_2022_CVPR}, and position embedding (PE)\cite{chu2021conditional,wu2021rethinking,shaw2018self,d2021convit}. However, seldom do they notice the way of joining PE to the network. To be more specific, most of the VTs add the PE directly to the patch embedding by default, and take them as the input of Transformer Encoders. 

\begin{figure}[t]
  \centering
  \includegraphics[width=1.0\linewidth]{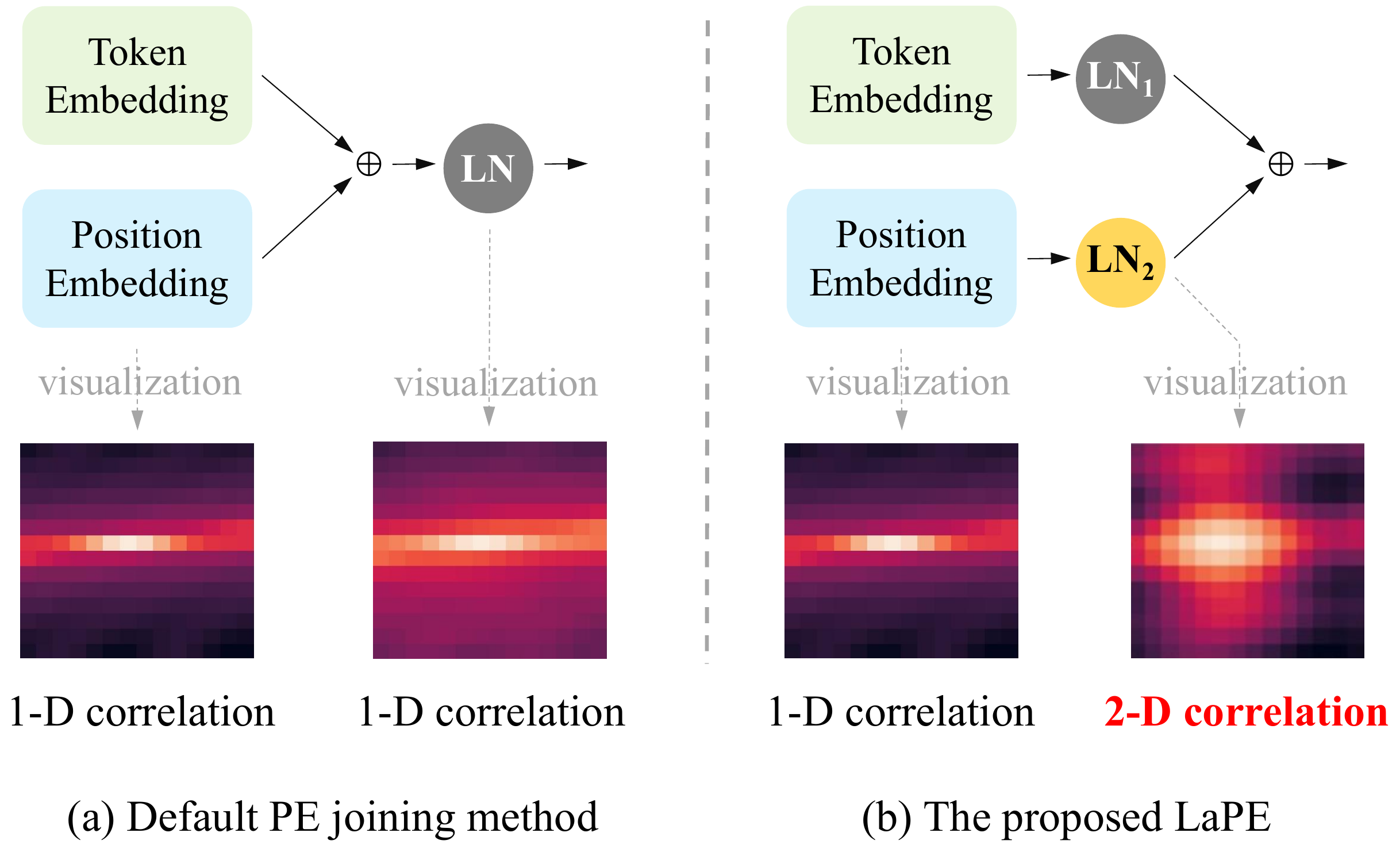}
   \caption{\textbf{A brief illustration of the default and proposed position embedding (PE) joining methods, with 1-D sinusoidal PE.} (a) By default, token embedding and position embedding are treated with the same layer normalization (LN). The PE still appears 1-D correlation after the LN operation. (b) We argue that token embedding and position embedding need separate LNs (LN$_1$,LN$_2$), as they need different affine transformation to adjust their expressiveness. We can see that the 1-D PE is transformed into 2-D correlation after operated by a separate LN.}
   \label{fig:1}
\end{figure}

In this paper, we analyze the input and output of each encoder layer in VTs using reparameterization and visualization, and find that the default PE joining method has inherent drawbacks, which limit the performance of VTs. The Layer Normalization (LN)\cite{ba2016layer} in VTs consists of per-token normalization and per-channel affine transformation. The affine transformation coefficients are learned to compensate for the possible loss of expressiveness caused by normalization \cite{wu2018group,ba2016layer}. Most of the VTs directly add PE to patch (token) embedding, then pass through an LN module, which means the PE and token embedding are operated by the same affine transformation. However, PE and token embedding are totally different information, so the affine transformation coefficients have to trade off between them, which limits the expressiveness of PE and hence constrains the performance of VTs. Fig.~\ref{fig:1}~(a) provides an illustration.

% We analyze the input and output of each layer in VTs by reparameterization and visualization. And we find that we can decouple the input of each layer into two kinds of embeddings. One is the summation of patch emebdding and token embeddings after each Multi-Head Self-Attention (MSA) and Muti-Layer Perceptron (MLP) block. The other is the PE, and its value and weight are same among layers. However, this will leads two main limitaions.

%仿射变换可以提升normalize之后的表达能力（cite），因此LN广泛应用于VTs。但由于默认方法直接将PE加到token embedding上，再做一个统一的LN,  这个LN的仿射变换就必须在PE和token embedding的表达能力上做权衡。从而限制PE的表达能力，进而影响VT的性能
%为了消除这种局限，同时最小化对网络的修改，最小化参数量和计算损耗，我们提出LaPE。。。。。
%实验表明，使用LaPE的VTs基本上没有额外的计算损耗和参数量，但是性能有明显提升，可视化表明，LaPE能显著提升PE的表示能力，比如在T2T中将1-D的sin PE通过仿射变换展现出2-D的位置correlation，比如在DeiT中将相对不变的PE通过仿射变换展现出了由local到global的hierarchical

To overcome this limitation with minimum cost on extra parameters and computational consumption, we propose to use two independent LNs for token embedding and PE for each layer, and add them together as the input of each Muti-Head Self-Attention (MSA) module, as shown in Fig.~\ref{fig:1}~(b). We name this new PE joining method Layer-adaptive Position Embedding (LaPE). Unlike many other works focusing on designing new PEs for VTs \cite{bello2019attention,wu2021rethinking}, LaPE focuses on the PE joining method, which is in parallel and compatible with these works. LaPE can be applied to learnable and sinusoidal absolute PE, and even relative PE, with stable performance improvement. Such a simple modification can significantly enhance the expressiveness of PE, like transforming a 1-D sinusoidal PE into 2-D one, see Fig.~\ref{fig:1} and Fig.~\ref{fig:3}. Moreover, it allows the model to adaptively adjust PE for each layer, like yielding hierarchical PEs that change from local to global as the layer goes deeper, see Fig.~\ref{fig:2}.

Extensive experiments on classification tasks demonstrate that LaPE is an effective and robust method that can improve various VTs with different PE types on multiple datasets. For VTs with learnable absolute PE, LaPE improves \textbf{0.94\%} accuracy for ViT-Lite\cite{hassani2021escaping} on Cifar10\cite{krizhevsky2009learning}, \textbf{0.98\%} for CCT\cite{hassani2021escaping} on Cifar100\cite{krizhevsky2009learning}, and \textbf{1.72\%} for DeiT-Ti\cite{touvron2021training} on ImageNet\cite{deng2009imagenet}. What's more, LaPE improves \textbf{0.19\%} accuracy for T2T-ViT-7\cite{yuan2021tokens} with 1-D sinusoidal PE and \textbf{0.30\%} for Swin-Transformer\cite{liu2021swin} with relative PE on ImageNet.
LaPE can also make VTs robust to PE types. Original DeiT-Ti\cite{touvron2021training} shows a performance gap of \textbf{3.84\%} between sinusoidal PE (67.70\%) and learnable PE (71.54\%). However, LaPE further improves the performance of DeiT with sinusoidal PE (72.22\% increased by 4.52\%) and learnable PE (73.26\% increase by 1.72\%), and shrinks the gap to \textbf{1.04\%}. This is remarkable considering the negligible extra parameters, memory and computational cost brought by LaPE.

To conclude, our contribution includes:
\begin{itemize}
  \item[1] We provide theoretical analysis on the default use of PE in common VTs and reveal its limitations.  %理论分析了default的缺陷，有可视化依据
  \item[2] We propose the LaPE, a new PE joining method, which is easy to implement and deploy. We reveal that LaPE can improve the expressiveness of PE and elevate the model performance. 
  \item[3] We verify that LaPE is effective and robust to various VTs with different PE types on multiple datasets, through extensive experiments.
%   \item[4] We reveal that LaPE greatly alleviates the problem that VTs are sensitive to PE types, which benefits the design of VTs.
  % LaPE可以使得模型对于PE type鲁棒，进而减少模型设计者在PE类型上的尝试。
\end{itemize}

% To overcome these limitations, we propose to add the layer normalized PE ahead of the MSA module for each layer. By doing this, we can provide VTs with layer-adpative position information, and we call this method Layer-adaptive Position Embedding (LaPE). 
% (1) PE and token embeddding has different affine transformation.
% (2) per-channel affine transformation coefficients can adjust the position information to meet the needs of one Transformer layer.
% (3) The coefficients of LN are different among layers, which results in hierarchical position information.

%------------------------------------------------------------------------
%------------------------------------------------------------------------
\begin{figure*}[t]
  \centering
%   \fbox{\rule{0pt}{0.5in} \rule{0.9\linewidth}{0pt}}
  \includegraphics[width=1.0\linewidth]{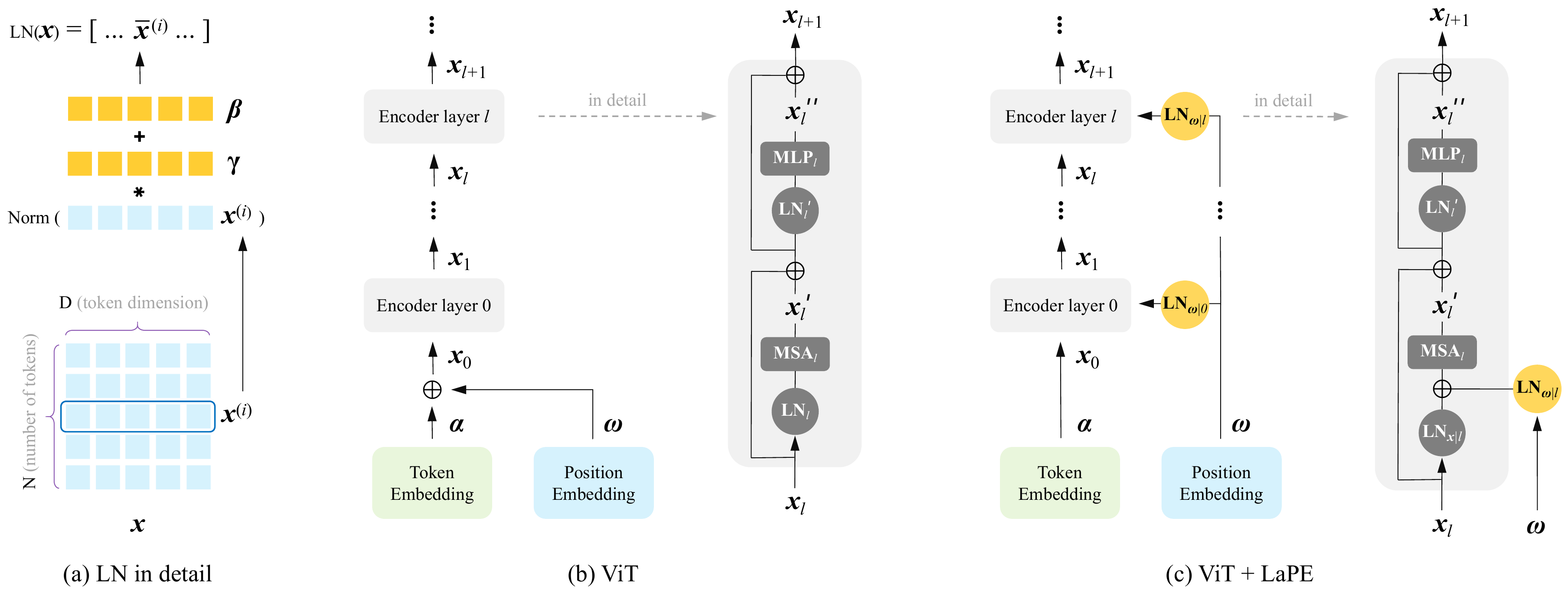}
   \caption{\textbf{Illustrations.} (a) Details of layer normalization (LN). (b) Typical ViT\cite{dosovitskiy2020image} structures \textit{(left)}, with detailed illustration of a encoder layer\textit{(right)}. (c) Apply LaPE to ViT. Specifically, we add independent LNs for PE at eacy layer, and add it to the layer normalized token embedding as the input of MSA module.}
   \label{fig:2}
\end{figure*}

\section{Related Work}
%------------------------------------------------------------------------
\subsection{Vision Transformers}
Transformer was originally introduced for natural language processing  \cite{vaswani2017attention}, and recently extended to computer vision tasks, including image classification\cite{touvron2021training,yuan2021tokens,liu2021swin,dosovitskiy2020image}, detection\cite{carion2020end,He_2022_CVPR2,Chen_2022_CVPR,Chen_2021_CVPR}, segmentation\cite{yuan2019segmentation,Xu_2022_CVPR,Hu_2021_CVPR,strudel2021segmenter}, 3D\cite{Wang_2022_CVPR,He_2022_CVPR,Fan_2022_CVPR,Li_2021_CVPR}, and cross-modal tasks\cite{jin2022expectationmaximization,li2022toward}, etc. Since we validate our method on image classification task, we summarize its representative works. ViT\cite{dosovitskiy2020image} is the first pure transformer for image classification, after which Vision Transformer (VT) becomes a research highlight. T2T-ViT\cite{yuan2021tokens} improves the tokenization part. DeiT\cite{touvron2021training} adds a distillation token. PVT\cite{wang2021pyramid} and PiT\cite{heo2021rethinking} adopt hierarchical structure. CvT\cite{wu2021cvt} and CeiT\cite{yuan2021incorporating} use the convolution to provide VT with inductive bias. Swin-Transformer\cite{liu2021swin,Liu_2022_CVPR} use the window attention. These VTs all use absolute or relative position embedding (PE). However, seldom do they notice the limitations of existing PE joining method.

%------------------------------------------------------------------------
\subsection{Position Embedding}
Since the self-attention mechanism is permutation-equivalent\cite{vaswani2017attention,dosovitskiy2020image}, Vision Transformer (VT) needs PE to identify tokens from different positions. The PE can either be fixed or learnable, absolute or relative.

\paragraph{Absolute Position Embedding.} The absolute PE encodes each position to distinguish tokens. It is usually added to the patch embedding before entering the Transformer encoders. In the original Transformer\cite{vaswani2017attention} and ViT\cite{dosovitskiy2020image}, the PE is generated by the fixed sinusoidal functions of different frequencies. The sinusoidal functions are designed to provide PE with locally monotonous similarity, so that PE can make VTs pay more attention to tokens close to each other\cite{wang2020position}. The sinusoidal PE in Transformer\cite{vaswani2017attention} and ViT\cite{dosovitskiy2020image} is 1-D, which can sense the sequence length. Meanwhile, there are 2-D sinusoidal PE\cite{wang2021translating,raisi20212lspe}, which has image height and width sensing. Moreover, the absolute PE can also be learnable, which is randomly initialized and updated with model's parameters. 
% SHAPE\cite{kiyono2021shape} further improves the shift-invariance of absolute position embedding.
%Except these commonly used absolute PE, there are many follow-up works [cite paper] proposing other absolute PE for VTs.

\paragraph{Relative Position Embedding.} The relative PE encodes the relative position between each pair of tokens. It first assigns a unique code to each relative position, and then involves the relative position embedding in the attention calculation. For natural language processing, the relative PE is first proposed in \cite{shaw2018self}, then further improved in XL-Net\cite{yang2019xlnet}, T5\cite{raffel2020exploring} and DeBERTa\cite{he2020deberta}. For vision tasks, a 2-D relative position embedding is firstly proposed in \cite{bello2019attention}, which is also used in Swin-Transformer\cite{liu2021swin}. iRPE\cite{wu2021rethinking} further improves the 2-D relative position embedding in its index function and relative position calculation. It is worth mentioning that our method is compatible with these works, and can further improve their performance.

\subsection{Position Information Fusing Modules}
There are some works \cite{d2021convit,chu2021conditional} arguing that VTs do not need explicit PE. Instead, they design position information fusing modules to provide VTs with implicit position information. ConViT\cite{d2021convit} proposes a Gated Positional Self-Attention module to balance learning content-based attention and position-based attention. CPVT\cite{chu2021conditional} proposes a convolution-based Positional Encoding Generator module, which generates position information for token embedding. These works have some limitations compared with our method. Firstly, they all tend to modify the model and propose new pipelines, thus they are inconvenient to transplant to other models. Secondly, these newly designed modules bring obvious extra computation and parameters. In contrast, our proposed LaPE is a PE joining method universal to all VTs, and the increased parameters, memory and computation consumption are negligible, while the performance gains are obvious.

%------------------------------------------------------------------------
%------------------------------------------------------------------------
\section{Method}
In this section, we first provide some preliminary knowledge about the layer normalization (LN) and the use of PE in Vision Transformers (VTs). Then we provide theoretical analysis on the default use of PE in common VTs and elaborate on the proposed LaPE. Next, we analyze the proposed LaPE and the default PE joining method by visualization. Finally, we show the implementation details on how to apply LaPE to general VTs.

\subsection{Preliminaries}
\paragraph{Layer Normalization.}Let us review the Layer Normalization (LN)\cite{ba2016layer}. Given a target tensor $\textbf{\textit{x}}\in \mathbb{R}^{ N\times D}$ that consists of $N$ tokens $\textbf{\textit{x}}^{(i)}\in \mathbb{R}^{1\times D}$, the operation of $\text{LN}(\textbf{\textit{x}})$ normalizes each token and applies channel-wise affine transformations, which can be formulated as:
\begin{equation}
   \begin{split}\label{E000}
        \Bar{\textbf{\textit{x}}}^{(i)}&=\boldsymbol{\gamma}\ast\frac{\textbf{\textit{x}}^{(i)}-\text{E}[\textbf{\textit{x}}^{(i)}]}{\sqrt{\text{Var}[\textbf{\textit{x}}^{(i)}]+\boldsymbol{\epsilon}}}+\boldsymbol{\beta},\\
        \text{LN}(\textbf{\textit{x}})&=[\Bar{\textbf{\textit{x}}}^{(1)}, ..., \Bar{\textbf{\textit{x}}}^{(N)}],
   \end{split}
\end{equation}
where $\text{E}[\textbf{\textit{x}}^{(i)}]$ and $\text{Var}[\textbf{\textit{x}}^{(i)}]$ represent the mean and variance of $\textbf{\textit{x}}^{(i)}$. $\boldsymbol{\gamma}\in \mathbb{R}^{1\times D}$ and $\boldsymbol{\beta}\in \mathbb{R}^{1\times D}$ represent the trainable affine transformation coefficients. Operator $\ast$ denotes element-wise multiplication. $\boldsymbol{\epsilon}$ is a small constant for division stability. Fig.~\ref{fig:2}~(a) illustrates the process of Eq.~(\ref{E000}).

\paragraph{The Use of PE in Vision Transformers.}
The core framework of typical Vision Transformers (VTs) consists of series encoder layers. The input of the first layer is:
\begin{equation}
   \begin{split}\label{E1}
       &\textbf{\textit{x}}_0=\boldsymbol{\alpha}+\boldsymbol{\omega},
   \end{split}
\end{equation}
where $\boldsymbol{\alpha}$ and $\boldsymbol{\omega}$ represent the token embedding and the PE, respectively. The following process of each layer can be formulated as:
\begin{equation}
\label{eq:x_l'}
       {\textbf{\textit{x}}_l}^{\prime}=\text{MSA}_{l}(\text{LN}_{l}(\textbf{\textit{x}}_l)),
\end{equation}
\begin{equation}
       {\textbf{\textit{x}}_l}^{\prime\prime}=\text{MLP}_{l}(\text{LN}_{l^{\prime}}(\textbf{\textit{x}}_l+{\textbf{\textit{x}}_l}^{\prime})),\\
\end{equation}
\begin{equation}
\label{E2}
       \textbf{\textit{x}}_{l+1}=\textbf{\textit{x}}_l+{\textbf{\textit{x}}_l}^{\prime}+{\textbf{\textit{x}}_l}^{\prime\prime},\\
\end{equation}
where $l$ is the index of layer, MSA denotes the Multi-Head Self-Attention module, MLP denotes the Multi-Layer Perceptron. $\text{LN}_l$ and $\text{LN}_{l^{\prime}}$ represent different LN module before MSA and MLP. Fig.~\ref{fig:2}~(b) illustrates these processes.

\begin{figure}[t]
  \centering
  \includegraphics[width=1.0\linewidth]{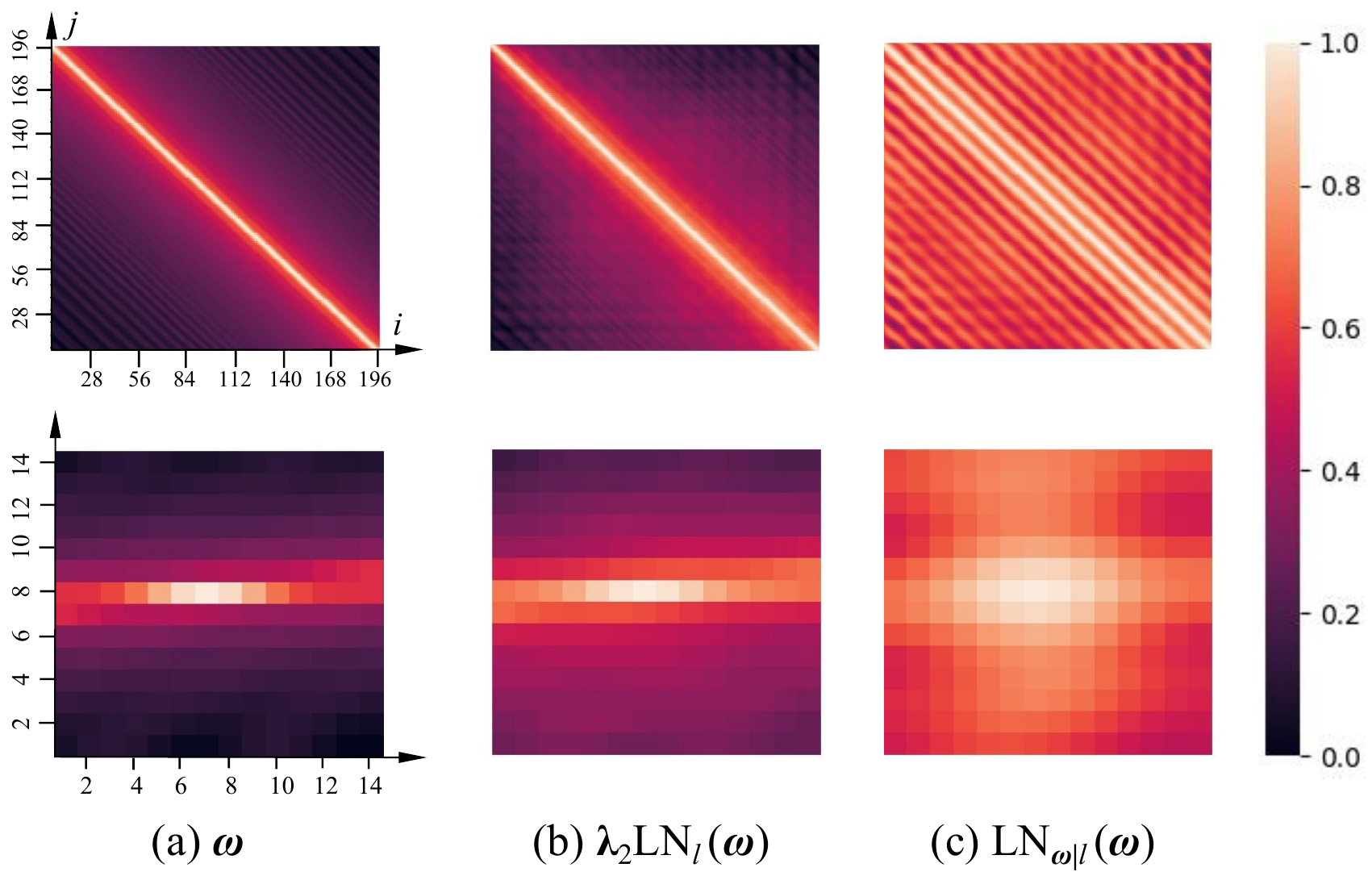}
   \caption{Visualization of the position correlations. (a) The original 1-D sinusoidal PE $\boldsymbol{\omega}$ shows 1-D position correlations. (b) $\boldsymbol{\lambda_2}\text{LN}_l(\boldsymbol{\omega})$ in Eq.~(\ref{eq:x_l' repara}) exhibits limited 2-D correlations. (c) $\text{LN}_{\boldsymbol{\omega}|l}(\boldsymbol{\omega})$ shows significant 2-D correlations.}
%   The first row visualizes the cross correlation matrix of different position representations. The second row visualizes the correlation between the 91th token and other tokens. }
   \label{fig:3}
\end{figure}

\subsection{LaPE for Vision Transformers}
Intuitively, the position embedding (PE) $\boldsymbol{\omega}$ added to the first layer can propagate to deeper layers due to the skip connections. By reparameterize $\textbf{\textit{x}}_l$ (see Appendix 1 for detailed derivation), we can rewrite Eq.~(\ref{eq:x_l'}) as:
\begin{equation}
\label{eq:x_l' repara}
    \begin{split}
       {\textbf{\textit{x}}_l}^{\prime} & =\text{MSA}_{l}(\text{LN}_{l}(\boldsymbol{\alpha}+\boldsymbol{\omega}+\sum\limits_{k=0}\limits^{l-1}(\textbf{\textit{x}}_{k}^{\prime}+\textbf{\textit{x}}_{k}^{\prime\prime}))),\\
       & =\text{MSA}_{l}(\text{LN}_{l}(\tilde{\textbf{\textit{x}}}+\boldsymbol{\omega}))\\
    %   & =\text{MSA}(\frac{\sigma_{\mathbf{\tilde{x}}}}{\sigma_{\mathbf{\tilde{x}}+\boldsymbol{\omega}}}\text{LN}_{l}(\tilde{x})+\frac{\sigma_{\boldsymbol{\omega}}}{\sigma_{\mathbf{\tilde{x}}+\boldsymbol{\omega}}}\text{LN}_{l}(\boldsymbol{\omega})+\frac{\sigma_{\mathbf{\tilde{x}}+\boldsymbol{\omega}}-\sigma_{\mathbf{\tilde{x}}}-\sigma_{\boldsymbol{\omega}}}{\sigma_{\mathbf{\tilde{x}}+\boldsymbol{\omega}}}\boldsymbol{\beta})\\
       & =\text{MSA}_{l}(\boldsymbol{\lambda_1}\text{LN}_{l}(\tilde{\textbf{\textit{x}}})+\boldsymbol{\lambda_2}\text{LN}_{l}(\boldsymbol{\omega})+\boldsymbol{\lambda_3}\boldsymbol{\beta}_l),\\
\end{split}
\end{equation}
where we use $\tilde{\textbf{\textit{x}}}$ to represent $\boldsymbol{\alpha}+\sum\limits_{k=0}\limits^{l-1}(\textbf{\textit{x}}_{k}^{\prime}+\textbf{\textit{x}}_{k}^{\prime\prime})$ then split $\text{LN}_{l}(\tilde{\textbf{\textit{x}}}+\boldsymbol{\omega})$ into three parts. $\boldsymbol{\lambda}\in \mathbb{R}^{N\times 1}$ represent token-wise coefficients, with following values:
\begin{equation}
\label{eq:lambda}
    \begin{split}
       \boldsymbol{\lambda_1} &=\frac{\boldsymbol{\sigma}_{\tilde{\textbf{\textit{x}}}}}{\boldsymbol{\sigma}_{\tilde{\textbf{\textit{x}}}+\boldsymbol{\omega}}},\\
       \boldsymbol{\lambda_2} &=\frac{\boldsymbol{\sigma}_{\boldsymbol{\omega}}}{\boldsymbol{\sigma}_{\tilde{\textbf{\textit{x}}}+\boldsymbol{\omega}}},\\
       \boldsymbol{\lambda_3} &=\frac{\boldsymbol{\sigma}_{\tilde{\textbf{\textit{x}}}+\boldsymbol{\omega}}-\boldsymbol{\sigma}_{\tilde{\textbf{\textit{x}}}}-\boldsymbol{\sigma}_{\boldsymbol{\omega}}}{\boldsymbol{\sigma}_{\tilde{\textbf{\textit{x}}}+\boldsymbol{\omega}}},
      \end{split}
\end{equation}
where $\boldsymbol{\sigma}_{(\cdot)}\in \mathbb{R}^{N\times 1}$ is the token-wise standard deviation.
% where we can clearly see that each layer's input involves a fixed PE $\boldsymbol{\omega}$. However, due to the serial process of transformer layers in VT, the importance of position information to different layers may not be equal. But the fixed PE for each layer does not consider the potential needs for different position information.

% In equation\ref{eq:x_l' repara}, we split the values in ${LN}_{l}$ into three parts, token embeddings $\tilde{x}$, PE $\boldsymbol{\omega}$ and bias. 

From Eq.~(\ref{eq:x_l' repara}), we can see that the token embeddings $\tilde{\textbf{\textit{x}}}$ share the same affine transformation coefficients with the position embedding $\boldsymbol{\omega}$. The affine transformation in LN is to compensate for the loss of expressiveness caused by normalization \cite{ba2016layer,wu2018group}. However, when token and position embedding are coupled, the affine transformation coefficients have to trade-off between these two embeddings, limiting the expressiveness of token embeddings and PE (see Fig.~\ref{fig:1}). 

To overcome this limitation with minimum cost on extra parameters and computational consumption, we propose to use two independent LNs for token embeddings and PE for each layer, and add them together as the input of each layer's MSA module. This allows the model to independently and adaptively adjust the expressiveness of PE for different layers.

Specifically, we set the input of the first layer as 
\begin{equation}
   \label{LaPE_E1}
       \textbf{\textit{x}}_0=\boldsymbol{\alpha},
\end{equation}
then modify Eq.~(\ref{eq:x_l'}) into:
\begin{equation}
\label{eq:lnpe x_l'}
       {\textbf{\textit{x}}_l}^{\prime}=\text{MSA}_{l}(\text{LN}_{\textbf{\textit{x}}|l}(\textbf{\textit{x}}_l)+\text{LN}_{\boldsymbol{\omega}|l}(\boldsymbol{\omega})).
\end{equation}
Note that $\text{LN}_{\textbf{\textit{x}}|l}$ and $\text{LN}_{\boldsymbol{\omega}|l}$ own different affine transformation coefficients. Fig.~\ref{fig:2}~(c) illustrates this modification.

We name the critical operation in Eq.~(\ref{eq:lnpe x_l'}) the \textbf{L}ayer-\textbf{a}daptive \textbf{P}osition \textbf{E}mbedding ({\bf LaPE}), which is robust and effective to diverse VTs and PE types. 

\begin{figure}[t]
  \centering
  \includegraphics[width=1.0\linewidth]{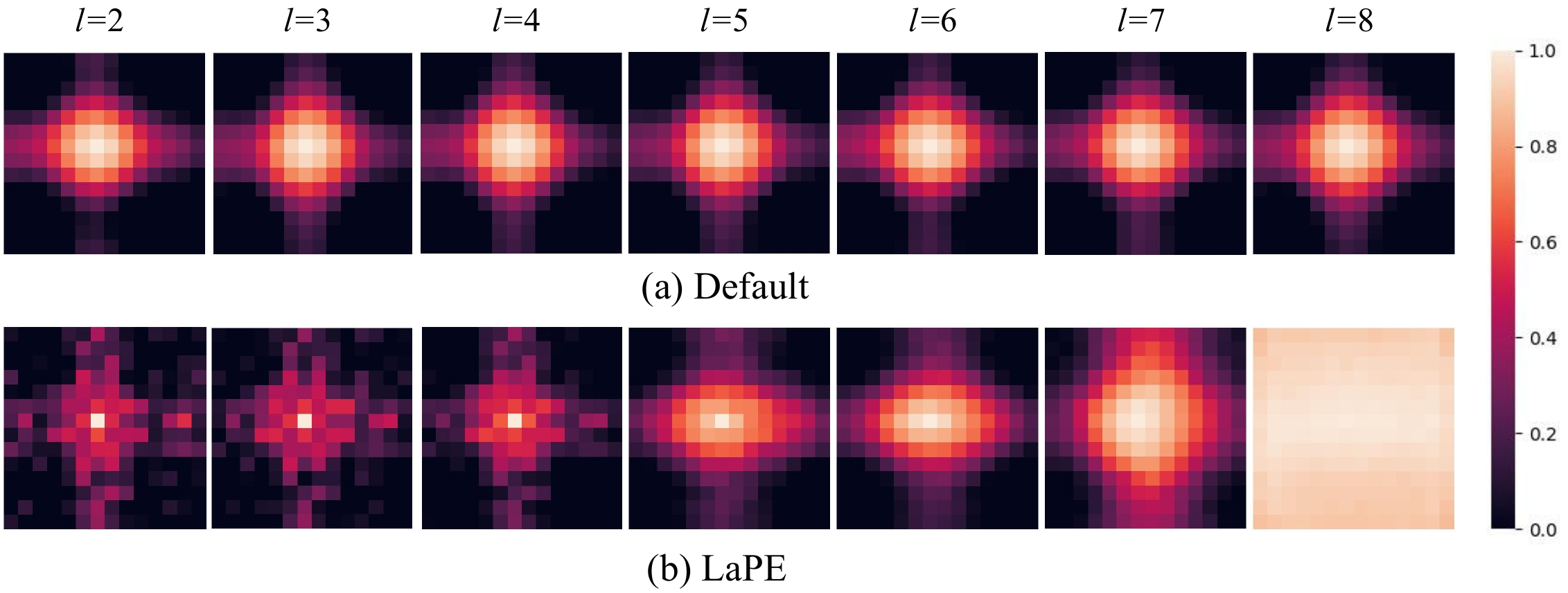}
   \caption{Visualization of the position correlations at different layers. (a) The default position correlation seems monotonic among different layers. (b) LaPE-based position correlation changes from local to global as the layer goes deeper.}
   \label{fig:4}
\end{figure}

\subsection{Analyzing LaPE Qualitatively}
We provide visualization for the position correlations of PEs, which is the position information contained in PEs\cite{wang2020position11,wang2020position22,dufter2022position}. The visualization results strongly support our analysis: (1) Using the same LN for token and position embeddings limits the position expressiveness; (2) Using two independent LN for token and position embeddings improves position expressiveness. For example, LaPE can transform 1-D PE into 2-D PE, and transform monotonic PEs into hierarchical ones.

\paragraph{Implementation of Visualization.}The PE in VTs describes the position correlations between each token. The position correlation can be measured by the cosine similarity between each token's PE:
\begin{equation}
\label{E3}
    s_{i,j} = \frac{\boldsymbol{\omega}^{(i)} \boldsymbol{\omega}^{(j)T}}{||\boldsymbol{\omega}^{(i)}||~ ||\boldsymbol{\omega}^{(j)}||},
\end{equation}
where $\boldsymbol{\omega}^{(i)}\in \mathbb{R}^{1\times D}$ and $\boldsymbol{\omega}^{(j)}\in \mathbb{R}^{1\times D}$ denotes the $i$th and $j$th token's PE, respectively. $s_{i,j}$ represents the position correlation between the $i$th and $j$th token.

\paragraph{From 1-D to 2-D.}We choose T2T-ViT-7\cite{yuan2021tokens} for demonstration since it uses 1-D sinusoidal PE. For the original T2T-ViT-7, we visualize every $s_{i,j}$ by converting $s_{i,j}$ into color pixels and combining all pixels into an image, as is shown in the upper part of Fig.~\ref{fig:3}~(a), where the horizontal and vertical axes denote the token index $i$ and $j$. 

Since the original tokens are taken from 2-D images, so we reshape the position correlations into 2-D for better intuitive observation. Specifically, we reshape the 96th row (since it is the center of the image) in the upper part of Fig.~\ref{fig:3}~(a) into a 2-D heat map (with shape 14$\times$14), as shown in the lower part of Fig.~\ref{fig:3}~(a). Now each value corresponds exactly to the token position of input image. In this way, we can intuitively see the relationship between the token's position correlations and their spatial positions.

%这样可以更直观地看出token之间的位置相关性与实际空间位置之间的关系。

From Fig.~\ref{fig:3}~(b), we can clearly see that the position correlation is 1-D. This is understandable since T2T-ViT uses the 1-D sinusoidal PE, which only has horizontal position perception and can not sense vertical position. To visualize the position correlation in T2T-ViT-7 with and without LaPE, we choose the 2nd layer ($l$=2), and calculate its cosine similarity $s_{i,j}$ for $\boldsymbol{\lambda_2}\text{LN}_l(\boldsymbol{\omega})$ in Eq.~(\ref{eq:x_l' repara}) (default PE joining method) and $\text{LN}_{\boldsymbol{\omega}|l}(\boldsymbol{\omega})$ in Eq.~(\ref{eq:lnpe x_l'}) (LaPE), respectively. This yields Fig.~\ref{fig:3}~(b) and Fig.~\ref{fig:3}(c). We can clearly see that Fig.~\ref{fig:3}~(b) shows limited 2-D position correlations, while Fig.~\ref{fig:3}~(c) shows evident 2-D position correlations, indicating that the positional expressiveness is significantly improved by the independent normalization and affine transformations.  
% \begin{equation}
% \label{E3}
%     s_{i,j} = \frac{\text{LN}_{l}(\boldsymbol{\omega}^{(i)}) \text{LN}_{l}(\boldsymbol{\omega}^{(j)T})}{||\text{LN}_{l}(\boldsymbol{\omega}^{(i)})||~||\text{LN}_{l}(\boldsymbol{\omega}^{(j)})||},
% \end{equation}
% and use the same visualization method mentioned above, yielding Fig.~\ref{fig:3}(c) and Fig.~\ref{fig:3}(d). 

\paragraph{From Monotonic to Hierarchical.} Here we choose DeiT-Ti\cite{touvron2021training} for example. From the 2nd layer to the 8th layer, we visualize the position correlations with the mentioned visualization method. Fig.~\ref{fig:4}~(a) shows the visualization of $\boldsymbol{\lambda_2}\text{LN}_l(\boldsymbol{\omega})$ in Eq.~(\ref{eq:x_l' repara}) (from DeiT-Ti with default PE), where the position correlations seem monotonic among layers. Fig.~\ref{fig:4}~(b) shows the visualization of $\text{LN}_{\boldsymbol{\omega}|l}(\boldsymbol{\omega})$ in Eq.~(\ref{eq:lnpe x_l'}) (from DeiT-Ti with LaPE), where the position correlations change obviously from local to global as the layer goes deeper, and the classification accuracy improves by 1.72\%. This phenomenon well fits our intuition that VTs may process information in a hierarchical way, thus they need hierarchical position information, and LaPE makes it possible.

\subsection{Appling LaPE to VTs}

We introduce a hyperparameter $\eta$ into LaPE, representing the number of layers that use LaPE. Through visualizing each layer's position correlation of $\text{LN}_{\boldsymbol{\omega}|l}(\boldsymbol{\omega})$ in Eq.~(\ref{eq:x_l' repara}), we find that the position correlations of the later layers are usually too global, which means the position information makes no difference. As shown in Fig.~\ref{fig:4}~(b), the LaPE-based position correlation of DeiT-Ti changes obviously from local to global as the layer goes deeper, and the figure of the last layer is nearly all white. By conducting experiments on various VTs, we find that using LaPE for each layer can improve the performance generally, but may not reach the \textit{optimal} performance for a few models. For example, LaPE achieves better performance by adding to the first 3 layers ($\eta=3$) of T2T-ViT-7\cite{yuan2021tokens} (with 7 layers in total) and the first 6 layers ($\eta=6$) of CVT\cite{hassani2021escaping} (with 7 layers in total). Unless otherwise stated, we use LaPE for all layers by default, as it is easy to implement and usually achieves good performance. 

For absolute PE, the LaPE is added before entering the MSA module for each layer. For relative PE, the LaPE is added to the Query-Key product as a position bias in the MSA module for each layer. The newly added parameters are the affine transformation coefficients of LNs for PE, and they are learned and updated with the model. See Appendix 2 for detailed pseudo codes of implementation. Furthermore, the newly added parameters are insignificant compared with the parameters of the model. For example, the amount of newly added parameters is 4.6k (joining LaPE each layer) for DeiT-Ti with 5M parameters. Meanwhile, the increased time and memory consumption are also negligible, as shown in Tab.~\ref{tab:4.7}.

Since the PE and model parameters are fixed during inference, we can pre-calculate the layer normalized PEs, i.e., $\text{LN}_{\boldsymbol{\omega}|l}(\boldsymbol{\omega})$ of Eq.~(\ref{eq:x_l' repara}), and use them directly when testing different images. This strategy can reduce the repetitive LN calculations in PE processing. In this way, LaPE increases almost no time and negligible memory consumption during inference, which is verified in Tab.~\ref{tab:4.7}.

%------------------------------------------------------------------------
%------------------------------------------------------------------------

\begin{table*}[t]
    \setlength{\tabcolsep}{10pt}
    \centering
    \small
    \begin{tabular}{c|c|c|c|cc}
        \toprule[1.2pt]
        \multirow{2}{*}{Model} & \multirow{2}{*}{Architecture} & \multirow{2}{*}{PE type} & \multirow{2}{*}{PE joining method} & \multicolumn{2}{c}{ImageNet Top1}\\
        \cline{5-6}
        &  & &  & 100 epoch & 300 epoch\\
        \toprule[1.2pt]
        \multirow{2}{*}{DeiT-Ti\cite{touvron2021training}} & \multirow{6}{*}{Pure Transformer} & \multirow{4}*{Learnable} & \rule{0pt}{9pt}Default & 58.13 & 71.54\\
        & & & \rule{0pt}{9pt}\textbf{LaPE} & \textbf{60.96} & \textbf{73.26}\\
        % \cline{4-6}
        \multirow{2}{*}{DeiT-S\cite{touvron2021training}} &  & & \rule{0pt}{9pt}Default & 68.41 & 80.00\\
        & & & \rule{0pt}{9pt}\textbf{LaPE} & \textbf{69.24} & \textbf{80.27}\\
        \cline{3-6}
        \multirow{2}{*}{T2T-ViT-7$^{\ast}$ \cite{yuan2021tokens}} & &  \multirow{2}*{Sinusoidal} & \rule{0pt}{9pt}Default & 65.62 & 71.69\\
        & & & \rule{0pt}{9pt}\textbf{LaPE} & \textbf{66.05} & \textbf{71.88}\\
        % \cline{4-6}
        % \multirow{2}{*}{T2T-ViT-10$^{\ast}$\cite{yuan2021tokens}} &  & & Default & 69.17 & 75.20\\
        % & & & \textbf{LaPE} & 69.21 & \\
        \hline
        \multirow{2}{*}{DeiT-Ti-distill\cite{touvron2021training}} & \multirow{4}{*}{Transformer with Distillation} & \multirow{4}*{Learnable} & \rule{0pt}{9pt}Default & 61.89 & 74.16\\
        & & & \rule{0pt}{9pt}\textbf{LaPE} & \textbf{63.38} & \textbf{75.06}\\
        % \cline{4-6}
        \multirow{2}{*}{DeiT-S-distll\cite{touvron2021training}} &  & & \rule{0pt}{9pt}Default & 70.65 & 80.98\\
        & & & \rule{0pt}{9pt}\textbf{LaPE} & \textbf{71.29} & \textbf{81.27}\\
        \hline
        % \multirow{2}{*}{PiT-Ti} & \multirow{8}{*}{\makecell[c]{Transformer with \\Hierarchical Structure}} & Default &  \multirow{8}*{learnable} &  & \\
        % & & LaPE & &  & \\
        % \multirow{2}{*}{PiT-S} &  & Default &  & & \\
        % & & LaPE & & &\\
    %   \multirow{2}{*}{PVT-Ti} & \multirow{4}{*}{\makecell[c]{Transformer with \\Hierarchical Structure}} & PE each layer & \multirow{4}*{learnable} &  & \\
    %     & & LaPE & &  & \\
    %     \multirow{2}{*}{PVT-S} &  & Default &  & & \\
    %     & & LaPE & & &\\
    %     \hline
        \multirow{2}{*}{Swin-Ti\cite{liu2021swin}} & \multirow{4}{*}{\makecell[c]{Transformer with \\window-based self-attention}} & \multirow{4}*{RPE} & \rule{0pt}{9pt}Default & 73.56 & 81.13\\
        & & & \rule{0pt}{9pt}\textbf{LaPE} & \textbf{73.82} & \textbf{81.18}\\
        \multirow{2}{*}{Swin-S\cite{liu2021swin}} &  & & \rule{0pt}{9pt}Default & 75.48 & 82.68\\
        & & & \rule{0pt}{9pt}\textbf{LaPE} & \textbf{76.50} & \textbf{82.98}\\
        \hline
        \multirow{2}{*}{CeiT-Ti\cite{yuan2021incorporating}} & \multirow{4}{*}{\makecell[c]{Transformer with \\Convolutional Inductive Bias}} & \multirow{4}*{Learnable} & \rule{0pt}{9pt}Default & 66.91 & 76.52\\
        & & & \rule{0pt}{9pt}\textbf{LaPE} & \textbf{67.09} & \textbf{76.67}\\
        \multirow{2}{*}{CeiT-S\cite{yuan2021incorporating}} &  & & \rule{0pt}{9pt}Default &  73.60 & 81.88 \\
        & & & \rule{0pt}{9pt}\textbf{LaPE} & \textbf{73.80} & \textbf{82.08}\\
        
    	\bottomrule[1.2pt]
    \end{tabular}
\caption{\textbf{Results on ImageNet-1K.} As shown here, applying LaPE to VTs improves their performance and accelerates the convergence on ImageNet-1K. LaPE is effective and robust to VTs with different architectures and different PE types. $\ast$ means using LaPE for partial layers.}
\label{tab:4.1}
\end{table*}
\begin{table*}[t]
    \centering
    \small
    \begin{tabular}{c|c|c|c|cc}
        \toprule[1.2pt]
        \multirow{2}{*}{Model} & \multirow{2}{*}{Architecture}& \multirow{2}{*}{PE type} & \multirow{2}{*}{PE joining method} & \multicolumn{2}{c}{Top1 Acc.}\\
        \cline{5-6}
        &  & &  & CIFAR-10 Top1 & CIFAR-100 Top1\\
        \toprule[1.2pt]
        \multirow{2}{*}{ViT-Lite\cite{hassani2021escaping}} & \multirow{2}{*}{\makecell[c]{Pure Transformer}} &  \multirow{2}*{learnable} & \rule{0pt}{9pt}Default & 93.448 & 74.984\\
        & &  & \rule{0pt}{9pt}\textbf{LaPE} & \textbf{94.386} & \textbf{75.424}\\
        \hline
        \multirow{2}{*}{CVT\cite{hassani2021escaping}} & \multirow{2}{*}{\makecell[c]{Transformer with \\Sequence Pooling}} & \multirow{2}*{learnable} & \rule{0pt}{9pt}Default & 94.302 & 77.452\\
        & & & \rule{0pt}{9pt}\textbf{LaPE} & \textbf{94.624} & \textbf{77.940}\\
        \hline
        \multirow{2}{*}{CCT\cite{hassani2021escaping}} & \multirow{2}{*}{\makecell[c]{Transformer with \\Convolutional Inductive Bias}} & \multirow{2}*{learnable} & \rule{0pt}{9pt}Default & 96.034 & 80.928\\
        & & & \rule{0pt}{9pt}\textbf{LaPE} & \textbf{96.474} & \textbf{81.904} \\
    	\bottomrule[1.2pt]
    \end{tabular}
\caption{\textbf{Results on CIFAR-10 and CIFAR-100.} As shown here, LaPE can further improve the performance of VTs that are specially designed for tiny datasets. It is worth noticing that the performance on CIFAR-10 is saturated (reaching around 95\%), while LaPE can still bring obvious improvement to all these VTs.}
\label{tab:4.2}
\end{table*}

\section{Experiments}
In this section, we conduct experiments to verify the effectiveness of the proposed LaPE on image classification task. Firstly, we choose various VTs and datasets, and evaluate LaPE with them. Then we analyze the consumption brought by LaPE to illustrate its efficiency. Afterward, we compare LaPE with other PE joining methods. Finally, we conduct extensive ablation studies on LaPE.
%------------------------------------------------------------------------
\subsection{Verifying LaPE on Representative VTs}
\paragraph{Datasets.} We evaluate our method on small and medium size datasets. For small size datasets, we evaluate VTs (with and without LaPE) on CIFAR-10 and CIFAR-100\cite{krizhevsky2009learning} with 50K training samples and 10K testing samples for 10 classes and 100 classes, respectively. For middle size dataset, we conduct experiments on ILSVRC-2012 ImageNet\cite{deng2009imagenet} with 1281K training samples and 50K testing samples for 1K classes.

\paragraph{Models.} To verify the robustness and generalizability of LaPE to different kinds of models and PE types on various datasets, we choose some representative VTs specially designed for small datasets (CIFAR-10\cite{krizhevsky2009learning} and CIFAR-100\cite{krizhevsky2009learning}) and medium dataset (ImageNet-1K\cite{deng2009imagenet}). On small datasets, we conduct experiments with ViT-Lite\cite{hassani2021escaping} (pure Transformer), CVT\cite{hassani2021escaping} (Transformer with sequence pooling) and CCT\cite{hassani2021escaping} (Transformer with convolutional inductive bias). These three models all use the learnable absolute PE. On medium dataset, we conduct experiments with DeiT\cite{touvron2021training} using learnable absolute PE and T2T-ViT\cite{yuan2021tokens} using 1-D sinusoidal absolute PE (pure Transformer); DeiT-distill using learnable absolute PE (Transformer with distillation); Swin\cite{liu2021swin} using 2-D relative PE (Transformer with window-based self-attention); CeiT\cite{yuan2021incorporating} using learnable absolute PE (Transformer with convolutional inductive bias).  We select two variants for DeiT, Swin, and CeiT: tiny and small, represented by Ti and S, respectively. We select T2T-ViT with the depth of 7, denoted as T2T-ViT-7. We choose ViT-Lite and CVT with the depth of 7 and kernel size of 4, and CCT with the depth of 7, kernel size of 3, and convolution layer of 1.

\paragraph{Implementation Details.} For fair comparison, we use the same settings as those in the original papers for models with and without LaPE. Specifically, all VTs are trained for 300 epochs with 224$\times$224 resolution images on ImageNet\cite{deng2009imagenet}, and with 32$\times$32 resolution images on CIFAR-10 and CIFAR-100\cite{krizhevsky2009learning}. For experiments on CIFAR-10 and CIFAR-100, we run 5 rounds with different random seeds (121, 122, 123, 124, 125) and use the averages as the final results. All VTs are trained on a single node with 1 (on CIFAR) or 4 (on ImageNet) V100 GPUs.

\paragraph{Results.} We conduct experiments on representative VTs mentioned above using LaPE and default PE joining method on ImageNet-1K\cite{deng2009imagenet}, and the results are shown in Tab.~\ref{tab:4.1}. According to the results, we find that LaPE can bring improvement to different VTs. Since DeiT and DeiT-distill have less local information, LaPE can bring obvious improvement to them. As VTs with window-based self-attention (Swin) and convolutional inductive bias (CeiT) already have strong locality information, the performance gains to them are not as obvious as to DeiT. We train Swin with 4 GPUs, different from 8 GPUs in the original paper, so its basic results (81.13 for Swin-Ti, 82.68 for Swin-S) are slightly lower than those in the original paper (81.20 for Swin-Ti, 83.20 for Swin-S). It is worth noticing that LaPE significantly accelerates the convergence, as can be observed from the accuracy at 100 epochs. Fig.~\ref{fig:5} shows the convergence curves of DeiT-Ti. 

We also conduct experiments on CIFAR-10 and CIFAR-100\cite{krizhevsky2009learning}. As shown in Tab.~\ref{tab:4.2}, we can see that LaPE can even bring 0.4\%$\sim$0.9\% gains of accuracy to saturated performance on CIFAR-10, and brings 0.4\%$\sim$1.0\% gains of accuracy on CIFAR-100. It is worth noting that the PE is optional for  CCT\cite{hassani2021escaping} with default PE joining method, as whether using PE yields comparable results. However, for CCT with LaPE, using PE can bring 0.44\% and 0.98\% performance gains on CIFAR-10 and CIFAR-100, respectively, implying that LaPE really improves the expressiveness of PE and further improves the classification performance. 

\begin{table}[t]
    \centering
    \scalebox{0.8}{
    \begin{tabular}{c|c|c|c|c}
        \toprule[1.2pt]
        Model & Stage & PE joining method & \makecell{Memory\\(MB)} & \makecell{Time\\(s/epoch)}\\
        \toprule[1.2pt]
        \multirow{4}{*}{DeiT-Ti} & \multirow{2}{*}{training} &\rule{0pt}{10pt}  Default & 10799 & 680\\
        & & \rule{0pt}{10pt} LaPE & 10822 & 699 \\
        \cline{2-5}
        & \multirow{2}{*}{inference} & \rule{0pt}{10pt} Default & 2676 & 100\\
        & & \rule{0pt}{10pt} LaPE & 2762 & 101\\
        \bottomrule[1.2pt]
    \end{tabular}}
\caption{\textbf{Comparison between LaPE and the default PE joining method on memory and time consumption during training and inference}, with batch size 256 on 4 V100 GPU. LaPE increases negligible extra consumption.}
\label{tab:4.7}
\end{table}
%------------------------------------------------------------------------
\subsection{Memory \& Time Consumption}
We record the memory and time consumption of the default PE joining method and LaPE in the training and inference stage. As shown in Tab.~\ref{tab:4.7}, we can see LaPE increases little memory and time consumption during training, and negligible consumption during inference. %Since each image shares the same PE, hence in inference stage, we may also pre-calculate the layer normalized PE for each layer in advance and use it when dealing with different images.%In training stage, LaPE operates new added LN for PE in each layer, which results in the negligibly increased memory and time.

\begin{table}[t]
    \centering
    \scalebox{0.8}{
    \begin{tabular}{c|c|c|c}
        \toprule[1.2pt]
        Model & PE type & PE joining method & ImageNet Top1\\
        \toprule[1.2pt]
        \multirow{13}{*}{DeiT-Ti\cite{touvron2021training}} & \multirow{3}{*}{1-D sinusoidal} & \rule{0pt}{10pt}basic PE &67.70\\
        &  & \rule{0pt}{10pt}shared PE & 70.66 \\
        &  & \rule{0pt}{10pt}\textbf{LaPE}  & \textbf{72.22}\\
        \cline{2-4}
        & \multirow{3}{*}{2-D sinusoidal} & \rule{0pt}{10pt}basic PE & 71.46\\
        &  & \rule{0pt}{10pt}shared PE & 71.47\\
        &  & \rule{0pt}{10pt}\textbf{LaPE} & \textbf{72.49}\\
        \cline{2-4}
        & \multirow{4}{*}{learnable} & \rule{0pt}{10pt}basic PE &71.54\\
        &  & \rule{0pt}{10pt}shared PE &72.00\\
        &  & \rule{0pt}{10pt}unshared PE &71.90\\
        &  & \rule{0pt}{10pt}\textbf{LaPE} &\textbf{73.26}\\
        % \cline{2-4}
        % & \multirow{2}{*}{RPE} & \rule{0pt}{10pt}unshared PE & running\\
        % &  & \rule{0pt}{10pt}\textbf{LaPE} & 72.67\\
    	\bottomrule[1.2pt]
    \end{tabular}}
\caption{\textbf{Comparision between LaPE and other PE joining methods with DeiT-Ti on ImageNet.} With different PE types, LaPE all shows the best performance compared with other PE joining methods. Meanwhile, LaPE makes DeiT-Ti robust to PE type, as it greatly reduces the performance gaps caused by using different PE types.}
\label{tab:4.3}
\end{table}

\subsection{Comparing LaPE with Other PE Joining Methods}
To prove the effectiveness and robustness of LaPE, we conduct experiments on DeiT-Ti\cite{touvron2021training} with various PE types and different PE joining methods. We choose three kinds of PE, including 1-D sinusoidal, 2-D sinusoidal, and learnable PE. We also choose four kinds of PE joining methods, which are basic PE, shared PE, unshared PE and LaPE. The basic PE means the default PE joining method, which adds the PE to patch embedding before entering the Transformer encoders. The shared PE means adding the same PE to the token embedding before entering each encoder layer. Similarly, the unshared PE means adding the layer-distinct PE before each encoder layer. Meanwhile, LaPE means operating the layer-distinct LN for the same PE before entering each MSA module. For 1-D and 2-D sinusoidal PE, we conduct experiments with three PE joining methods except for unshared PE, owning to the fixed and unlearnable PE type. For learnable PE, we conduct experiments with all these four PE joining methods. 

\begin{figure}[t]
  \centering
  \includegraphics[width=1.0\linewidth]{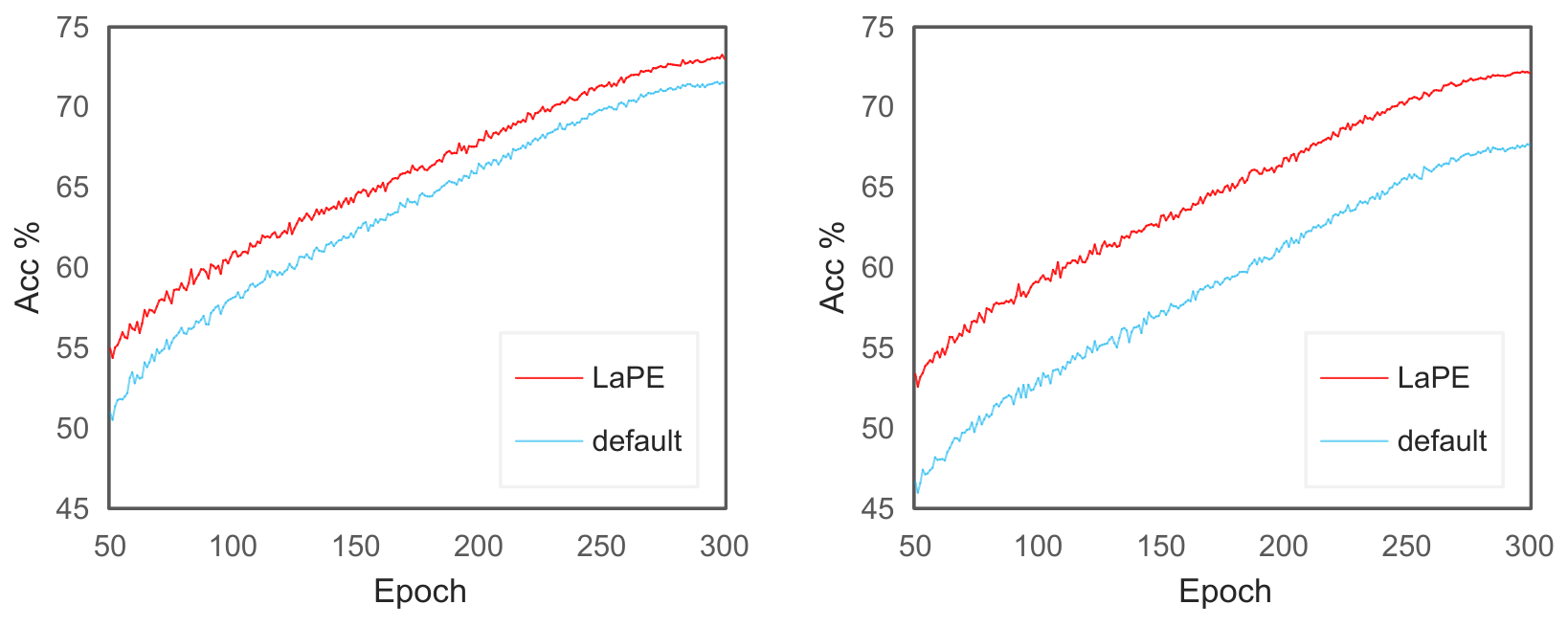}
   \caption{Convergence curve, default DeiT-Ti vs. LaPE-based DeiT-Ti. \textit{(left)} with learnable PE. \textit{(right)} with 1-D sinusoidal PE.}
   \label{fig:5}
\end{figure}

As shown in Tab.~\ref{tab:4.3}, LaPE achieves the best performance among each PE joining method. It is worth mentioning that LaPE even works better than unshared PE, which has more parameters to learn position information. This is because LaPE operates layer-distinct LN to the same PE for each layer, where the same PE prevents the model from overfitting the position information, and the layer-distinct LN learns to adaptively adjust the position information. Meanwhile, from the convergence curves in Fig.~\ref{fig:5}, we can see that LaPE alleviates the performance gap caused by different PE types, which means models with LaPE are robust to PE types. Thus, LaPE can improve efficiency when designing Transformer models. 
% as researchers no more need to select the appropriate PE type. 

% \begin{table}[t]
%     \centering
%     \scalebox{0.7}{
%     \begin{tabular}{c|c|c|c}
%         \toprule[1.2pt]
%         Model & PE joining method & Process of PE & ImageNet Top1\\
%         \toprule[1.2pt]
%         \multirow{7}{*}{DeiT-Ti} & Default & basic & \\
%         \cline{2-4}
%         & \multirow{6}{*}{Function(x) each layer} & per-channel gamma*norm(PE) & \\
%         &  & gamma*norm(PE) &  \\
%         &  & per-channel gamma*PE & \\
%         &  & gamma*PE & \\
%         &  & PE & \\
%         &  & LaPE & \\
%         \bottomrule[1.2pt]
%     \end{tabular}}
% \caption{\textbf{Ablation study on LaPE with Deit-Ti on ImageNet} .}
% \label{tab:4.4}
% \end{table}
% {\bf Models.}

%------------------------------------------------------------------------
\subsection{Ablation Study}
% To defend the design options in our method, 
In this section, we perform ablation studies on the proposed LaPE with ViT-Lite\cite{hassani2021escaping} on CIFAR-100. We first try to gradually remove each component in LaPE. Then we try applying LaPE to different layers. 

\paragraph{Decompose $\text{LN}_{\boldsymbol{\omega}|l}$.} As shown in Tab.~\ref{tab:4.5}, we conduct experiments on different components of $\text{LN}_{\boldsymbol{\omega}|l}$, based on ViT-Lite\cite{hassani2021escaping}. The Default configuration means the original ViT-Lite. The rest configurations all take the similar network structures as ViT-Lite + LaPE, which is shown in Fig.~\ref{fig:2}~(c), except for $\text{LN}_{\boldsymbol{\omega}|l}(\boldsymbol{\omega})$. In Tab.~\ref{tab:4.5}, the configuration $\boldsymbol{\omega}$ means replacing $\text{LN}_{\boldsymbol{\omega}|l}(\boldsymbol{\omega})$ in Eq.~(\ref{eq:lnpe x_l'}) with $\boldsymbol{\omega}$; $\gamma\boldsymbol{\omega}$ means replacing it with $\gamma\boldsymbol{\omega}$, where $\gamma$ is a scalar; $\boldsymbol{\gamma}\ast\boldsymbol{\omega}$ means replacing it with $\boldsymbol{\gamma}\ast\boldsymbol{\omega}$, where $\boldsymbol{\gamma}$ denotes a per-channel scale factor; $\boldsymbol{\gamma}\ast\boldsymbol{\omega}+\boldsymbol{\beta}$ means replacing it with $\boldsymbol{\gamma}\ast\boldsymbol{\omega}+\boldsymbol{\beta}$, where $\boldsymbol{\beta}$ denotes a per-channel bias. Norm$(\boldsymbol{\omega})$ means replacing it with Norm$(\boldsymbol{\omega})$, where Norm$(\boldsymbol{\omega})$ means operate per-token normalization to $\boldsymbol{\omega}$. So on and so forth. The final configuration $\boldsymbol{\gamma}\ast$Norm$(\boldsymbol{\omega})+\boldsymbol{\beta}$ is exactly $\text{LN}_{\boldsymbol{\omega}|l}(\boldsymbol{\omega})$.  

% The process of $\boldsymbol{\omega}$, $\gamma\boldsymbol{\omega}$, $\boldsymbol{\gamma}\ast \boldsymbol{\omega}$ and $\boldsymbol{\gamma}\ast\boldsymbol{\omega}+\boldsymbol{\beta}$ means adding PE, PE multiplied with a weight scalar, PE multiplied with a per-channel weight vector and PE with a affine transformation for each Transformer layer. The process of $\text{Norm(\boldsymbol{\omega})}$, $\gamma \text{Norm(\boldsymbol{\omega})}$,  $\boldsymbol{\gamma}\ast \text{Norm(\boldsymbol{\omega})}$ and $\boldsymbol{\gamma}\ast \text{Norm(\boldsymbol{\omega})}+\boldsymbol{\beta}$ mean adding normalized PE, normalized PE with a weight scalar, normalized PE with a per-channel weight vector and normalized PE with an affine transformation (layer normalized PE, LaPE) for each Transformer layer. 

The results in Tab.~\ref{tab:4.5} shows that the former four configurations, i.e., $\boldsymbol{\omega}$, $\gamma \boldsymbol{\omega}$, $\boldsymbol{\gamma}\ast \boldsymbol{\omega}$, and $\boldsymbol{\gamma}\ast \boldsymbol{\omega}+\boldsymbol{\beta}$ perform slightly lower than the default configuration. This is understandable since the un-normalized PE may deviate a lot from a normalized token embedding. The latter four configurations, i.e., Norm $(\boldsymbol{\omega})$, $\gamma$Norm$(\boldsymbol{\omega})$,  $\boldsymbol{\gamma}\ast$Norm$(\boldsymbol{\omega})$, and $\boldsymbol{\gamma}\ast$Norm$(\boldsymbol{\omega})+\boldsymbol{\beta}$ all perform better than the the default. Therefore, an independent normalization for PE is critical. However, we can see that $\gamma$Norm$(\boldsymbol{\omega})$ and  $\boldsymbol{\gamma}\ast$Norm$(\boldsymbol{\omega})$ yield worse results than Norm$(\boldsymbol{\omega})$, which means an intact affine transformation is crucial for normalized PE. In all, LaPE shows the best performance by comparison. 

\paragraph{LaPE for Partial Layers.} As shown in Tab.~\ref{tab:4.6}, we apply LaPE to different encoder layers in ViT-Lite\cite{hassani2021escaping}. For example, $\eta=4$ means we apply independent $\text{LN}_{\boldsymbol{\omega}|l}$ for PE at the 1st, 2nd, 3rd, and 4th layer, leaving the other layers unconnected (as default). The results show that adding independent $\text{LN}_{\boldsymbol{\omega}|l}$ for PE at all layers may not be the optimal choice, so the results in Tab.~\ref{tab:4.1} and Tab.~\ref{tab:4.2} have the potential to be improved since we simply apply LaPE for all layers for those models (except for T2T-ViT\cite{yuan2021tokens}).

\begin{table}[t]
    \centering
    \small
    \begin{tabular}{c|c|c}
        \toprule[1.2pt]
        Model & Configuration & CIFAR-100 Top1\\
        \toprule[1.2pt]
        \multirow{9}{*}{ViT-Lite\cite{hassani2021escaping}} & Default & 74.984\\
        & $\boldsymbol{\omega}$ & 74.084\\
        & $\gamma \boldsymbol{\omega}$ & 74.636\\
        & $\boldsymbol{\gamma}\ast \boldsymbol{\omega}$ & 74.518\\
        & $\boldsymbol{\gamma}\ast \boldsymbol{\omega}+\boldsymbol{\beta}$ & 74.250 \\
        & Norm($\boldsymbol{\omega}$) & 75.238\\
        & $\gamma$Norm($\boldsymbol{\omega}$) & 75.192\\
        & $\boldsymbol{\gamma}\ast$ Norm($\boldsymbol{\omega}$) & 74.952\\
        & $\boldsymbol{\gamma}\ast$
        \text{Norm($\boldsymbol{\omega}$)}+$\boldsymbol{\beta}$ & \textbf{75.424}\\
        \bottomrule[1.2pt]
    \end{tabular}
\caption{\textbf{Decompose LN$_{\boldsymbol{\omega}|l}$ in ViT-Lite}. $\boldsymbol{\omega}$ denotes the PE; $\gamma$ denotes the weight constant; $\boldsymbol{\gamma}$ accompanied by $\ast$ denotes per-channel weight vector; $\boldsymbol{\beta}$ denotes the per-channel bias; Norm$(\cdot)$ denotes the token-wise normalization. The results show that the standard LN$_{\boldsymbol{\omega}|l}$ (the last configuration) is the best choice.}
\label{tab:4.5}
\end{table}

\begin{table}[t]
    \centering
    \small
    \scalebox{1}{
    \begin{tabular}{c|c|c}
        \toprule[1.2pt]
        Model & PE joining method & CIFAR-100 Top1\\
        \toprule[1.2pt]
        \multirow{8}{*}{ViT-Lite\cite{hassani2021escaping}} & Default & 74.984\\
        \cline{2-3}
        & LaPE, $\eta=1$ & 74.062\\
        & LaPE, $\eta=2$ & 74.716\\
        & LaPE, $\eta=3$ & 75.468\\
        & LaPE, $\eta=4$ & 75.652\\
        & LaPE, $\eta=5$ & 75.658\\
        & LaPE, $\eta=6$ & \textbf{75.660}\\
        & LaPE, $\eta=7$ & 75.424\\
        \bottomrule[1.2pt]
    \end{tabular}}
\caption{\textbf{LaPE for partial layers.} As the joining layers increase, the performance keeps improving, except for the last layer. Starting from the third layer, the following configurations all perform better than the baseline.}
\label{tab:4.6}
\end{table}

%------------------------------------------------------------------------
%------------------------------------------------------------------------
%Specifically, LaPE uses two independent LNs for token embeddings and PE for each layer. In this way, LaPE can provide layer-adaptive and hierarchical position information for VTs., as LaPE brings obvious improvement to various VTs with different PE type on multi-size datasets.
\section{Conclusion \& Discussion}
In this paper, we study the position embedding (PE) in Vision Transformers (VTs), and propose a simple but effective method, LaPE. Specifically, LaPE uses two independent LNs for token embeddings and PE for each layer. In this way, LaPE can provide layer-adaptive and hierarchical position information for VTs. Extensive experiments and ablation studies demonstrate the superiority of our method. LaPE has potential to be an alternative PE joining method for general transformer-based models, and its effectiveness on dense predicted tasks deserves further study, as these tasks are more sensitive to position.
%, as LaPE brings obvious improvement to various VTs with different PE type on multi-size datasets

% \section{Limitation}
% 对于RPE的有效性值得进一步研究，对于一些具有inductive bias的模型提升不像纯transoformer那样明显。
Though with the mentioned merits, there are a few limitations of this method. For example, finding the optimal hyperparameter $\eta$ relies on experiment and lacks theoretical instruction. Though setting $\eta$ as full layers may not be the optimal, it usually yields good results.

{\small
\bibliographystyle{ieee_fullname}
\bibliography{egbib}

\begin{thebibliography}{10}\itemsep=-1pt

\bibitem{ba2016layer}
Jimmy~Lei Ba, Jamie~Ryan Kiros, and Geoffrey~E Hinton.
\newblock Layer normalization.
\newblock {\em arXiv preprint arXiv:1607.06450}, 2016.

\bibitem{bello2019attention}
Irwan Bello, Barret Zoph, Ashish Vaswani, Jonathon Shlens, and Quoc~V Le.
\newblock Attention augmented convolutional networks.
\newblock In {\em Proceedings of the IEEE/CVF international conference on
  computer vision}, pages 3286--3295, 2019.

\bibitem{carion2020end}
Nicolas Carion, Francisco Massa, Gabriel Synnaeve, Nicolas Usunier, Alexander
  Kirillov, and Sergey Zagoruyko.
\newblock End-to-end object detection with transformers.
\newblock In {\em European conference on computer vision}, pages 213--229.
  Springer, 2020.

\bibitem{Chen_2021_CVPR}
Ding-Jie Chen, He-Yen Hsieh, and Tyng-Luh Liu.
\newblock Adaptive image transformer for one-shot object detection.
\newblock In {\em Proceedings of the IEEE/CVF Conference on Computer Vision and
  Pattern Recognition (CVPR)}, pages 12247--12256, June 2021.

\bibitem{Chen_2022_CVPR}
Zhe Chen, Jing Zhang, and Dacheng Tao.
\newblock Recurrent glimpse-based decoder for detection with transformer.
\newblock In {\em Proceedings of the IEEE/CVF Conference on Computer Vision and
  Pattern Recognition (CVPR)}, pages 5260--5269, June 2022.

\bibitem{chu2021conditional}
Xiangxiang Chu, Zhi Tian, Bo Zhang, Xinlong Wang, Xiaolin Wei, Huaxia Xia, and
  Chunhua Shen.
\newblock Conditional positional encodings for vision transformers.
\newblock {\em Arxiv preprint 2102.10882}, 2021.

\bibitem{deng2009imagenet}
Jia Deng, Wei Dong, Richard Socher, Li-Jia Li, Kai Li, and Li Fei-Fei.
\newblock Imagenet: A large-scale hierarchical image database.
\newblock In {\em 2009 IEEE conference on computer vision and pattern
  recognition}, pages 248--255. Ieee, 2009.

\bibitem{dosovitskiy2020image}
Alexey Dosovitskiy, Lucas Beyer, Alexander Kolesnikov, Dirk Weissenborn,
  Xiaohua Zhai, Thomas Unterthiner, Mostafa Dehghani, Matthias Minderer, Georg
  Heigold, Sylvain Gelly, et~al.
\newblock An image is worth 16x16 words: Transformers for image recognition at
  scale.
\newblock {\em arXiv preprint arXiv:2010.11929}, 2020.

\bibitem{dufter2022position}
Philipp Dufter, Martin Schmitt, and Hinrich Sch{\"u}tze.
\newblock Position information in transformers: An overview.
\newblock {\em Computational Linguistics}, 48(3):733--763, 2022.

\bibitem{d2021convit}
St{\'e}phane d’Ascoli, Hugo Touvron, Matthew~L Leavitt, Ari~S Morcos, Giulio
  Biroli, and Levent Sagun.
\newblock Convit: Improving vision transformers with soft convolutional
  inductive biases.
\newblock In {\em International Conference on Machine Learning}, pages
  2286--2296. PMLR, 2021.

\bibitem{Fan_2022_CVPR}
Lue Fan, Ziqi Pang, Tianyuan Zhang, Yu-Xiong Wang, Hang Zhao, Feng Wang, Naiyan
  Wang, and Zhaoxiang Zhang.
\newblock Embracing single stride 3d object detector with sparse transformer.
\newblock In {\em Proceedings of the IEEE/CVF Conference on Computer Vision and
  Pattern Recognition (CVPR)}, pages 8458--8468, June 2022.

\bibitem{hassani2021escaping}
Ali Hassani, Steven Walton, Nikhil Shah, Abulikemu Abuduweili, Jiachen Li, and
  Humphrey Shi.
\newblock Escaping the big data paradigm with compact transformers.
\newblock {\em arXiv preprint arXiv:2104.05704}, 2021.

\bibitem{He_2022_CVPR}
Chenhang He, Ruihuang Li, Shuai Li, and Lei Zhang.
\newblock Voxel set transformer: A set-to-set approach to 3d object detection
  from point clouds.
\newblock In {\em Proceedings of the IEEE/CVF Conference on Computer Vision and
  Pattern Recognition (CVPR)}, pages 8417--8427, June 2022.

\bibitem{He_2022_CVPR2}
Liqiang He and Sinisa Todorovic.
\newblock Destr: Object detection with split transformer.
\newblock In {\em Proceedings of the IEEE/CVF Conference on Computer Vision and
  Pattern Recognition (CVPR)}, pages 9377--9386, June 2022.

\bibitem{he2020deberta}
Pengcheng He, Xiaodong Liu, Jianfeng Gao, and Weizhu Chen.
\newblock Deberta: Decoding-enhanced bert with disentangled attention.
\newblock {\em arXiv preprint arXiv:2006.03654}, 2020.

\bibitem{heo2021rethinking}
Byeongho Heo, Sangdoo Yun, Dongyoon Han, Sanghyuk Chun, Junsuk Choe, and
  Seong~Joon Oh.
\newblock Rethinking spatial dimensions of vision transformers.
\newblock In {\em Proceedings of the IEEE/CVF International Conference on
  Computer Vision}, pages 11936--11945, 2021.

\bibitem{Hu_2021_CVPR}
Li Hu, Peng Zhang, Bang Zhang, Pan Pan, Yinghui Xu, and Rong Jin.
\newblock Learning position and target consistency for memory-based video
  object segmentation.
\newblock In {\em Proceedings of the IEEE/CVF Conference on Computer Vision and
  Pattern Recognition (CVPR)}, pages 4144--4154, June 2021.

\bibitem{jin2022expectationmaximization}
Peng Jin, JinFa Huang, Fenglin Liu, Xian Wu, Shen Ge, Guoli Song, David~A.
  Clifton, and Jie Chen.
\newblock Expectation-maximization contrastive learning for compact
  video-and-language representations.
\newblock In Alice~H. Oh, Alekh Agarwal, Danielle Belgrave, and Kyunghyun Cho,
  editors, {\em Advances in Neural Information Processing Systems}, 2022.

\bibitem{krizhevsky2009learning}
Alex Krizhevsky, Geoffrey Hinton, et~al.
\newblock Learning multiple layers of features from tiny images.
\newblock 2009.

\bibitem{li2022toward}
Hao Li, Jinfa Huang, Peng Jin, Guoli Song, Qi Wu, and Jie Chen.
\newblock Toward 3d spatial reasoning for human-like text-based visual question
  answering.
\newblock {\em arXiv preprint arXiv:2209.10326}, 2022.

\bibitem{Li_2021_CVPR}
Ruibo Li, Guosheng Lin, Tong He, Fayao Liu, and Chunhua Shen.
\newblock Hcrf-flow: Scene flow from point clouds with continuous high-order
  crfs and position-aware flow embedding.
\newblock In {\em Proceedings of the IEEE/CVF Conference on Computer Vision and
  Pattern Recognition (CVPR)}, pages 364--373, June 2021.

\bibitem{Liu_2022_CVPR}
Ze Liu, Han Hu, Yutong Lin, Zhuliang Yao, Zhenda Xie, Yixuan Wei, Jia Ning, Yue
  Cao, Zheng Zhang, Li Dong, Furu Wei, and Baining Guo.
\newblock Swin transformer v2: Scaling up capacity and resolution.
\newblock In {\em Proceedings of the IEEE/CVF Conference on Computer Vision and
  Pattern Recognition (CVPR)}, pages 12009--12019, June 2022.

\bibitem{liu2021swin}
Ze Liu, Yutong Lin, Yue Cao, Han Hu, Yixuan Wei, Zheng Zhang, Stephen Lin, and
  Baining Guo.
\newblock Swin transformer: Hierarchical vision transformer using shifted
  windows.
\newblock In {\em Proceedings of the IEEE/CVF International Conference on
  Computer Vision}, pages 10012--10022, 2021.

\bibitem{loshchilov2017decoupled}
Ilya Loshchilov and Frank Hutter.
\newblock Decoupled weight decay regularization.
\newblock {\em arXiv preprint arXiv:1711.05101}, 2017.

\bibitem{Mao_2022_CVPR}
Xiaofeng Mao, Gege Qi, Yuefeng Chen, Xiaodan Li, Ranjie Duan, Shaokai Ye, Yuan
  He, and Hui Xue.
\newblock Towards robust vision transformer.
\newblock In {\em Proceedings of the IEEE/CVF Conference on Computer Vision and
  Pattern Recognition (CVPR)}, pages 12042--12051, June 2022.

\bibitem{raffel2020exploring}
Colin Raffel, Noam Shazeer, Adam Roberts, Katherine Lee, Sharan Narang, Michael
  Matena, Yanqi Zhou, Wei Li, Peter~J Liu, et~al.
\newblock Exploring the limits of transfer learning with a unified text-to-text
  transformer.
\newblock {\em J. Mach. Learn. Res.}, 21(140):1--67, 2020.

\bibitem{raisi20212lspe}
Zobeir Raisi, Mohamed~A Naiel, Georges Younes, Steven Wardell, and John Zelek.
\newblock 2lspe: 2d learnable sinusoidal positional encoding using transformer
  for scene text recognition.
\newblock In {\em 2021 18th Conference on Robots and Vision (CRV)}, pages
  119--126. IEEE, 2021.

\bibitem{rumelhart1986learning}
David~E Rumelhart, Geoffrey~E Hinton, and Ronald~J Williams.
\newblock Learning representations by back-propagating errors.
\newblock {\em nature}, 323(6088):533--536, 1986.

\bibitem{shaw2018self}
Peter Shaw, Jakob Uszkoreit, and Ashish Vaswani.
\newblock Self-attention with relative position representations.
\newblock {\em arXiv preprint arXiv:1803.02155}, 2018.

\bibitem{strudel2021segmenter}
Robin Strudel, Ricardo Garcia, Ivan Laptev, and Cordelia Schmid.
\newblock Segmenter: Transformer for semantic segmentation.
\newblock In {\em Proceedings of the IEEE/CVF International Conference on
  Computer Vision}, pages 7262--7272, 2021.

\bibitem{touvron2021training}
Hugo Touvron, Matthieu Cord, Matthijs Douze, Francisco Massa, Alexandre
  Sablayrolles, and Herv{\'e} J{\'e}gou.
\newblock Training data-efficient image transformers \& distillation through
  attention.
\newblock In {\em International Conference on Machine Learning}, pages
  10347--10357. PMLR, 2021.

\bibitem{vaswani2017attention}
Ashish Vaswani, Noam Shazeer, Niki Parmar, Jakob Uszkoreit, Llion Jones,
  Aidan~N Gomez, {\L}ukasz Kaiser, and Illia Polosukhin.
\newblock Attention is all you need.
\newblock {\em Advances in neural information processing systems}, 30, 2017.

\bibitem{wang2020position}
Benyou Wang, Lifeng Shang, Christina Lioma, Xin Jiang, Hao Yang, Qun Liu, and
  Jakob~Grue Simonsen.
\newblock On position embeddings in bert.
\newblock In {\em International Conference on Learning Representations}, 2020.

\bibitem{wang2020position22}
Benyou Wang, Lifeng Shang, Christina Lioma, Xin Jiang, Hao Yang, Qun Liu, and
  Jakob~Grue Simonsen.
\newblock On position embeddings in bert.
\newblock In {\em International Conference on Learning Representations}, 2020.

\bibitem{wang2021pyramid}
Wenhai Wang, Enze Xie, Xiang Li, Deng-Ping Fan, Kaitao Song, Ding Liang, Tong
  Lu, Ping Luo, and Ling Shao.
\newblock Pyramid vision transformer: A versatile backbone for dense prediction
  without convolutions.
\newblock In {\em Proceedings of the IEEE/CVF International Conference on
  Computer Vision}, pages 568--578, 2021.

\bibitem{Wang_2022_CVPR}
Yikai Wang, TengQi Ye, Lele Cao, Wenbing Huang, Fuchun Sun, Fengxiang He, and
  Dacheng Tao.
\newblock Bridged transformer for vision and point cloud 3d object detection.
\newblock In {\em Proceedings of the IEEE/CVF Conference on Computer Vision and
  Pattern Recognition (CVPR)}, pages 12114--12123, June 2022.

\bibitem{wang2020position11}
Yu-An Wang and Yun-Nung Chen.
\newblock What do position embeddings learn? an empirical study of pre-trained
  language model positional encoding.
\newblock {\em arXiv preprint arXiv:2010.04903}, 2020.

\bibitem{wang2022uformer}
Zhendong Wang, Xiaodong Cun, Jianmin Bao, Wengang Zhou, Jianzhuang Liu, and
  Houqiang Li.
\newblock Uformer: A general u-shaped transformer for image restoration.
\newblock In {\em Proceedings of the IEEE/CVF Conference on Computer Vision and
  Pattern Recognition}, pages 17683--17693, 2022.

\bibitem{wang2021translating}
Zelun Wang and Jyh-Charn Liu.
\newblock Translating math formula images to latex sequences using deep neural
  networks with sequence-level training.
\newblock {\em International Journal on Document Analysis and Recognition
  (IJDAR)}, 24(1):63--75, 2021.

\bibitem{wu2021cvt}
Haiping Wu, Bin Xiao, Noel Codella, Mengchen Liu, Xiyang Dai, Lu Yuan, and Lei
  Zhang.
\newblock Cvt: Introducing convolutions to vision transformers.
\newblock In {\em Proceedings of the IEEE/CVF International Conference on
  Computer Vision}, pages 22--31, 2021.

\bibitem{wu2021rethinking}
Kan Wu, Houwen Peng, Minghao Chen, Jianlong Fu, and Hongyang Chao.
\newblock Rethinking and improving relative position encoding for vision
  transformer.
\newblock In {\em Proceedings of the IEEE/CVF International Conference on
  Computer Vision}, pages 10033--10041, 2021.

\bibitem{wu2018group}
Yuxin Wu and Kaiming He.
\newblock Group normalization.
\newblock In {\em Proceedings of the European conference on computer vision
  (ECCV)}, pages 3--19, 2018.

\bibitem{Xu_2022_CVPR}
Lian Xu, Wanli Ouyang, Mohammed Bennamoun, Farid Boussaid, and Dan Xu.
\newblock Multi-class token transformer for weakly supervised semantic
  segmentation.
\newblock In {\em Proceedings of the IEEE/CVF Conference on Computer Vision and
  Pattern Recognition (CVPR)}, pages 4310--4319, June 2022.

\bibitem{yang2019xlnet}
Zhilin Yang, Zihang Dai, Yiming Yang, Jaime Carbonell, Russ~R Salakhutdinov,
  and Quoc~V Le.
\newblock Xlnet: Generalized autoregressive pretraining for language
  understanding.
\newblock {\em Advances in neural information processing systems}, 32, 2019.

\bibitem{yuan2021incorporating}
Kun Yuan, Shaopeng Guo, Ziwei Liu, Aojun Zhou, Fengwei Yu, and Wei Wu.
\newblock Incorporating convolution designs into visual transformers.
\newblock In {\em Proceedings of the IEEE/CVF International Conference on
  Computer Vision}, pages 579--588, 2021.

\bibitem{yuan2021tokens}
Li Yuan, Yunpeng Chen, Tao Wang, Weihao Yu, Yujun Shi, Zi-Hang Jiang,
  Francis~EH Tay, Jiashi Feng, and Shuicheng Yan.
\newblock Tokens-to-token vit: Training vision transformers from scratch on
  imagenet.
\newblock In {\em Proceedings of the IEEE/CVF International Conference on
  Computer Vision}, pages 558--567, 2021.

\bibitem{yuan2022volo}
Li Yuan, Qibin Hou, Zihang Jiang, Jiashi Feng, and Shuicheng Yan.
\newblock Volo: Vision outlooker for visual recognition.
\newblock {\em IEEE Transactions on Pattern Analysis and Machine Intelligence},
  2022.

\bibitem{yuan2019segmentation}
Yuhui Yuan, Xiaokang Chen, Xilin Chen, and Jingdong Wang.
\newblock Segmentation transformer: Object-contextual representations for
  semantic segmentation.
\newblock {\em arXiv preprint arXiv:1909.11065}, 2019.

\bibitem{zhang2022nested}
Zizhao Zhang, Han Zhang, Long Zhao, Ting Chen, Sercan~{\"O} Arik, and Tomas
  Pfister.
\newblock Nested hierarchical transformer: Towards accurate, data-efficient and
  interpretable visual understanding.
\newblock In {\em Proceedings of the AAAI Conference on Artificial
  Intelligence}, volume~36, pages 3417--3425, 2022.

\end{thebibliography}
}
% CVPR 2023 Paper Template
% based on the CVPR template provided by Ming-Ming Cheng (https://github.com/MCG-NKU/CVPR_Template)
% modified and extended by Stefan Roth (stefan.roth@NOSPAMtu-darmstadt.de)

%%%%%%%%% TITLE - PLEASE UPDATE
\clearpage
\onecolumn
\appendix
\begin{center}
    \Large \bf Supplementary Material for Position Embedding Needs \\ an Independent Layer Normalization \\ [10mm]
\end{center}

\section{Limitation of Position Information in Original Vision Transformer}
% \section{Decompose The Position Information in Each Encoder Layer}
In order to analyze the position information in each encoder layer, we reparameterize the input of each encoder and Multi-Head Self-Attention (MSA) module and find the limitation of the position information joined by default. In the original ViT\cite{dosovitskiy2020image}, the position embedding (PE) 
$\boldsymbol{\omega}$ is added to the patch embedding $\boldsymbol{\alpha}$ at the beginning and propagated to deeper layers due to the skip connections. The input of each encoder layer can be rewritten as:
\begin{equation}
\label{eq:x_l' repara}
    \begin{split}
       {\textbf{\textit{x}}_l} & = {\textbf{\textit{x}}_{l-1}} + \textbf{\textit{x}}_{l-1}^{\prime} + \textbf{\textit{x}}_{l-1}^{\prime\prime}\\
       & = {\textbf{\textit{x}}_{l-2}} + \textbf{\textit{x}}_{l-2}^{\prime} + \textbf{\textit{x}}_{l-2}^{\prime\prime}+ \textbf{\textit{x}}_{l-1}^{\prime} + \textbf{\textit{x}}_{l-1}^{\prime\prime}\\
       & = {\textbf{\textit{x}}_{0}}+\textbf{\textit{x}}_{0}^{\prime} + \textbf{\textit{x}}_{0}^{\prime\prime}+...+\textbf{\textit{x}}_{l-1}^{\prime} + \textbf{\textit{x}}_{l-1}^{\prime\prime}\\
       & =\boldsymbol{\alpha}+\boldsymbol{\omega}+\sum\limits_{k=0}\limits^{l-1}(\textbf{\textit{x}}_{k}^{\prime}+\textbf{\textit{x}}_{k}^{\prime\prime})\\
       & = \tilde{\textbf{\textit{x}}}_l+\boldsymbol{\omega},
    %   & =\text{MSA}(\frac{\sigma_{\mathbf{\tilde{x}}}}{\sigma_{\mathbf{\tilde{x}}+\boldsymbol{\omega}}}\text{LN}_{l}(\tilde{x})+\frac{\sigma_{\boldsymbol{\omega}}}{\sigma_{\mathbf{\tilde{x}}+\boldsymbol{\omega}}}\text{LN}_{l}(\boldsymbol{\omega})+\frac{\sigma_{\mathbf{\tilde{x}}+\boldsymbol{\omega}}-\sigma_{\mathbf{\tilde{x}}}-\sigma_{\boldsymbol{\omega}}}{\sigma_{\mathbf{\tilde{x}}+\boldsymbol{\omega}}}\boldsymbol{\beta})\\
\end{split}
\end{equation}
where $l$ is the index of layer, and $\textbf{\textit{x}}^{\prime}$, $\textbf{\textit{x}}^{\prime\prime} \in  \mathbb{R}^{N\times D}$ represent the output of each MSA and MLP module (refer to Fig.~2 of our paper for more details). We use $\tilde{\textbf{\textit{x}}}_l$ to represent $\boldsymbol{\alpha}+\sum\limits_{k=0}\limits^{l-1}(\textbf{\textit{x}}_{k}^{\prime}+\textbf{\textit{x}}_{k}^{\prime\prime})$. In this way, we separate the input of each layer into two parts, PE and token embeddings.

We can further rewrite the input of each MSA module.
\begin{equation}
\label{eq:MSA' repara}
    \begin{split}
       {\textbf{\textit{x}}_l}^{\prime} & =\text{MSA}_{l}(\text{LN}_{l}(\tilde{\textbf{\textit{x}}}_l+\boldsymbol{\omega}))\\
       & = \text{MSA}_{l}(\boldsymbol{\gamma}_l\ast\frac{\tilde{\textbf{\textit{x}}}_l+\boldsymbol{\omega}-\text{E}[\tilde{\textbf{\textit{x}}}_l+\boldsymbol{\omega}]}{\boldsymbol{\sigma}_{\tilde{\textbf{\textit{x}}}_l+\boldsymbol{\omega}}}+\boldsymbol{\beta}_l)\\
       & = \text{MSA}_{l}(\boldsymbol{\gamma}_l\ast\frac{\tilde{\textbf{\textit{x}}}_l-\text{E}[\tilde{\textbf{\textit{x}}}_l]}{\boldsymbol{\sigma}_{\tilde{\textbf{\textit{x}}}_l+\boldsymbol{\omega}}}+\boldsymbol{\gamma}_l\ast\frac{\boldsymbol{\omega}-\text{E}[\boldsymbol{\omega}]}{\boldsymbol{\sigma}_{\tilde{\textbf{\textit{x}}}_l+\boldsymbol{\omega}}}+\boldsymbol{\beta}_l)\\
       & = \text{MSA}_{l}((\frac{\boldsymbol{\sigma}_{\tilde{\textbf{\textit{x}}}_l}}{\boldsymbol{\sigma}_{\tilde{\textbf{\textit{x}}}_l+\boldsymbol{\omega}}}\boldsymbol{\gamma}_l\ast\frac{\tilde{\textbf{\textit{x}}}_l-\text{E}[\tilde{\textbf{\textit{x}}}_l]}{\boldsymbol{\sigma}_{\tilde{\textbf{\textit{x}}}_l}})+(\frac{\boldsymbol{\sigma}_{\boldsymbol{\omega}}}{\boldsymbol{\sigma}_{\tilde{\textbf{\textit{x}}}_l+\boldsymbol{\omega}}}\boldsymbol{\gamma}_l\ast\frac{\boldsymbol{\omega}-\text{E}[\boldsymbol{\omega}]}{\boldsymbol{\sigma}_{\boldsymbol{\omega}}})+\boldsymbol{\beta}_l)\\
       & = \text{MSA}_{l}(\frac{\boldsymbol{\sigma}_{\tilde{\textbf{\textit{x}}}_l}}{\boldsymbol{\sigma}_{\tilde{\textbf{\textit{x}}}_l+\boldsymbol{\omega}}}(\boldsymbol{\gamma}_l\ast\frac{\tilde{\textbf{\textit{x}}}_l-\text{E}[\tilde{\textbf{\textit{x}}}_l]}{\boldsymbol{\sigma}_{\tilde{\textbf{\textit{x}}}_l}}+\boldsymbol{\beta}_l)+\frac{\boldsymbol{\sigma}_{\boldsymbol{\omega}}}{\boldsymbol{\sigma}_{\tilde{\textbf{\textit{x}}}_l+\boldsymbol{\omega}}}(\boldsymbol{\gamma}_l\ast\frac{\boldsymbol{\omega}-\text{E}[\boldsymbol{\omega}]}{\boldsymbol{\sigma}_{\boldsymbol{\omega}}}+\boldsymbol{\beta}_l)+\frac{\boldsymbol{\sigma}_{\tilde{\textbf{\textit{x}}}_l+\boldsymbol{\omega}}-\boldsymbol{\sigma}_{\tilde{\textbf{\textit{x}}}_l}-\boldsymbol{\sigma}_{\boldsymbol{\omega}}}{\boldsymbol{\sigma}_{\tilde{\textbf{\textit{x}}}_l+\boldsymbol{\omega}}}\boldsymbol{\beta}_l)\\
       & = \text{MSA}_{l}(\frac{\boldsymbol{\sigma}_{\tilde{\textbf{\textit{x}}}_l}}{\boldsymbol{\sigma}_{\tilde{\textbf{\textit{x}}}_l+\boldsymbol{\omega}}}\text{LN}_{l}(\mathbf{\tilde{x})}+\frac{\boldsymbol{\sigma}_{\boldsymbol{\omega}}}{\boldsymbol{\sigma}_{\tilde{\textbf{\textit{x}}}_l+\boldsymbol{\omega}}} \text{LN}_{l}(\boldsymbol{\omega})+\frac{\boldsymbol{\sigma}_{\tilde{\textbf{\textit{x}}}_l+\boldsymbol{\omega}}-\boldsymbol{\sigma}_{\tilde{\textbf{\textit{x}}}_l}-\boldsymbol{\sigma}_{\boldsymbol{\omega}}}{\boldsymbol{\sigma}_{\tilde{\textbf{\textit{x}}}_l+\boldsymbol{\omega}}}\boldsymbol{\beta}_l)\\
    %   & =\text{MSA}(\frac{\sigma_{\mathbf{\tilde{x}}}}{\sigma_{\mathbf{\tilde{x}}+\boldsymbol{\omega}}}\text{LN}_{l}(\tilde{x})+\frac{\sigma_{\boldsymbol{\omega}}}{\sigma_{\mathbf{\tilde{x}}+\boldsymbol{\omega}}}\text{LN}_{l}(\boldsymbol{\omega})+\frac{\sigma_{\mathbf{\tilde{x}}+\boldsymbol{\omega}}-\sigma_{\mathbf{\tilde{x}}}-\sigma_{\boldsymbol{\omega}}}{\sigma_{\mathbf{\tilde{x}}+\boldsymbol{\omega}}}\boldsymbol{\beta}_l)\\
       & =\text{MSA}_{l}(\boldsymbol{\lambda_1}\text{LN}_{l}(\mathbf{\tilde{x})}+\boldsymbol{\lambda_2}\text{LN}_{l}(\boldsymbol{\omega})+\boldsymbol{\lambda_3}\boldsymbol{\beta}_l),\\
\end{split}
\end{equation}
where LN represents the layer normalization operation, MSA represents the Multi-Head Self-Attention module, $\boldsymbol{\gamma} \in \mathbb{R}^{1\times D}$ and $\boldsymbol{\beta} \in \mathbb{R}^{1\times D}$ are the trainable affine transformation coefficients in LN, E$(\cdot) \in \mathbb{R}^{N\times 1}$ and  $\boldsymbol{\sigma}_{(\cdot)}\in \mathbb{R}^{N\times 1}$ are the token-wise mean and standard deviation. We use $\boldsymbol{\lambda_1}$, $\boldsymbol{\lambda_2}$, $\boldsymbol{\lambda_3} \in \mathbb{R}^{N\times 1}$ to represent the coefficients of LN($\tilde{\textbf{\textit{x}}}_l$), LN($\boldsymbol{\omega}$), $\boldsymbol{\beta}$, respectively. In this way, we successfully decompose the position information $\boldsymbol{\lambda_2}\text{LN}(\boldsymbol{\omega})$ in each encoder layer (we ignore some minor couplings). 

From Eq.~\ref{eq:MSA' repara}, we can see that the position embedding $\boldsymbol{\omega}$ shares the same LN as the token embeddings $\tilde{\textbf{\textit{x}}}_l$. However, the position and token embeddings represent different information. When these two kinds of embeddings are coupled, the affine transformation coefficients of the LN have to trade off between them, limiting the expressiveness of token and position embeddings.

\section{PyTorch-Like Pseudo Code Implementation}
We provide PyTorch-like codes here for easier understanding and better reproducibility of our proposed LaPE. 

For absolute PE, Vision Transformers (VTs) add the PE to the patch embedding before entering Transformer encoders by default, while VTs with LaPE add the layer normalized PE before entering each MSA module. We take the framework of DeiT\cite{touvron2021training} for example to illustrate the implementation.
\label{pseudo code}
\lstset{ %
language=python,                % the language of the code
basicstyle=\small,           % the size of the fonts that are used for the code
numbers=left,                   % where to put the line-numbers
numberstyle=\tiny\color{gray},  % the style that is used for the line-numbers
stepnumber=2,                   % the step between two line-numbers. If it's 1, each line 
                                % will be numbered
numbersep=16pt,                  % how far the line-numbers are from the code
backgroundcolor=\color{white},      % choose the background color. You must add \usepackage{color}
showspaces=false,               % show spaces adding particular underscores
showstringspaces=false,         % underline spaces within strings
showtabs=false,                 % show tabs within strings adding particular underscores
frame=single,                   % adds a frame around the code
rulecolor=\color{black},        % if not set, the frame-color may be changed on line-breaks within not-black text (e.g. commens (green here))
tabsize=2,                      % sets default tabsize to 2 spaces
captionpos=b,                   % sets the caption-position to bottom
breaklines=true,                % sets automatic line breaking
breakatwhitespace=false,        % sets if automatic breaks should only happen at whitespace
title=\lstname,                 % show the filename of files included with \lstinputlisting;
                            % also try caption instead of title
keywordstyle=\color{blue},          % keyword style
commentstyle=\color{green},       % comment style
stringstyle=\color{orange},         % string literal style
escapeinside={\%*}{*)},            % if you want to add LaTeX within your code
morekeywords={*,...}               % if you want to add more keywords to the set
}
\begin{lstlisting}
# Attention: Multi-Head Self-Attention calculation
# MLP: Multi-Layer Perceptron: Linear+Gelu+Linear+Dropout
# Transformer Block with LaPE
def LaPE_Block(x, pos_embed):
    # Eq.(9) add the layer normalized PE before entering MSA module
    x = x + Attention(LayerNorm1(x)+LayerNorm2(pos_embed))
    x = x + MLP(LayerNorm3(x))
    return x
    
def VisionTransformer(x):
    x = Patch_embed(x) # get the patch embedding
    x = cat(cls_token, x) # concate the class token
    x = x # Eq.(8) only pass the patch embedding to encoders
    for _ in range(depth): 
        x = LaPE_Block(x,pos_embed) # pass through series encoders
    x = Head(x[:, 0]) # classification
    return x
\end{lstlisting}

Relative PE is independently learned for each layer and is added to the Query-Key product as a position bias in each MSA module. We propose to take the layer normalized relative PE as the position bias, and add it to the Query-Key product in MSA modules. As the only difference between LaPE and default relative PE is the added position bias in attention calculation, we display their attention parts as follows.
\begin{lstlisting}
# relative_position_bias: the results indexed from the learned relative position bias table
def RPE_Attention_LaPE():
    qkv = Linear(x) # linear x to a higher dimension
    q, k, v = qkv[0], qkv[1], qkv[2] # get Query, Key, Value
    attn = q @ k * scale # get Query-Key product
    attn[:,:,1:,1:] += LayerNorm(relative_position_bias) # add the layer normalized RPE
    x = softmax(attn) @ v # newly attentioned token embedding
    return x
\end{lstlisting}

\begin{figure}[t]
  \centering
  \includegraphics[width=1.0\linewidth]{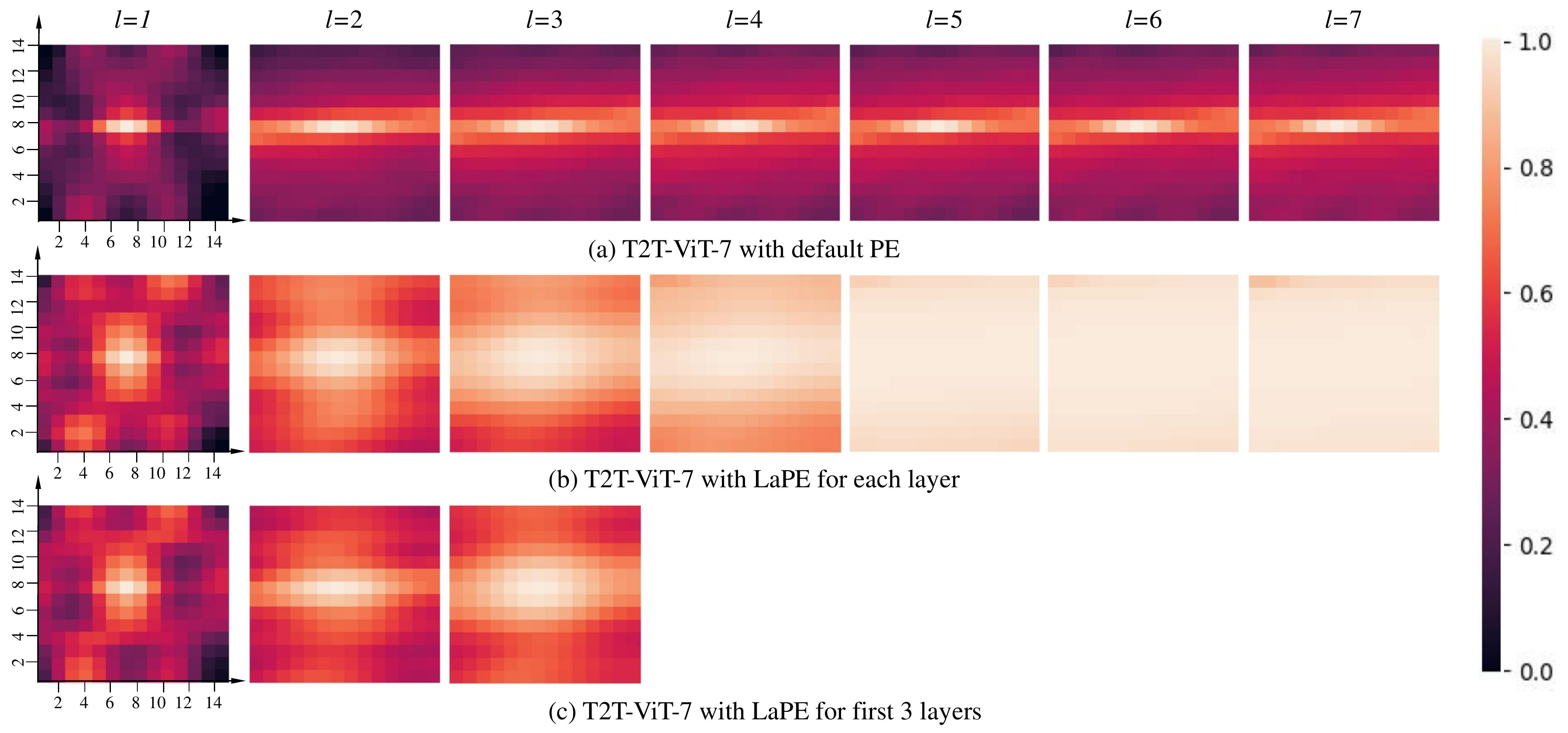}
   \caption{\textbf{Visualization of the position correlations at different layers for T2T-ViT-7.} (a) The default position correlations are 1-D and monotonic after the second layer. (b) The position correlations are 2-D and change from local to global. (c) The position correlations are 2-D and not completely global.}
   \label{fig:1}
\end{figure}

\begin{figure}[t]
  \centering
  \includegraphics[width=1.0\linewidth]{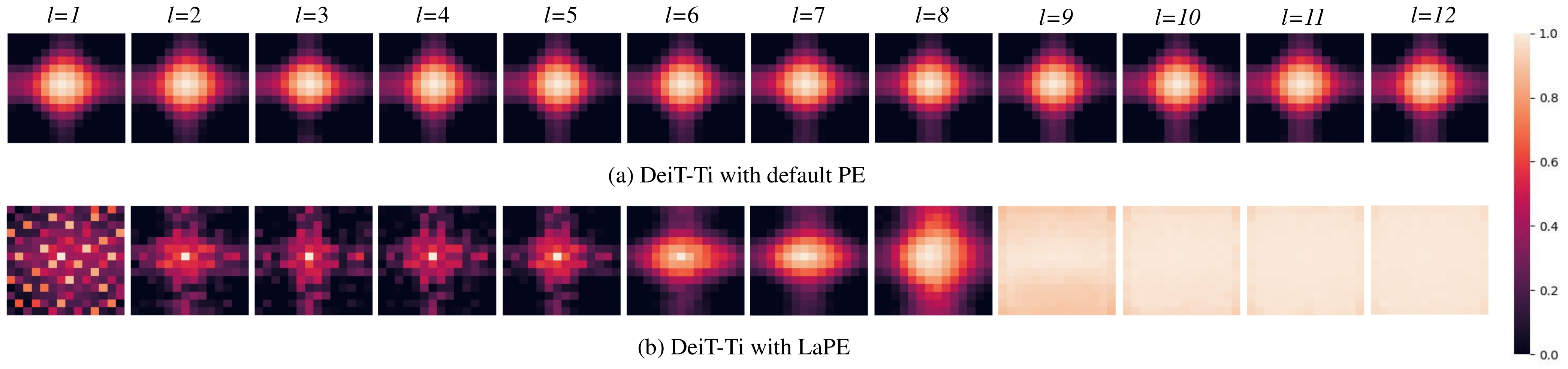}
   \caption{\textbf{Visualization of the position correlations at different layers for DeiT-Ti.} (a) The default position correlations are monotonic. (b) The position correlations change from local to global as the layer goes deeper.}
   \label{fig:2}
\end{figure}

\begin{figure}[t]
  \centering
  \includegraphics[width=1.0\linewidth]{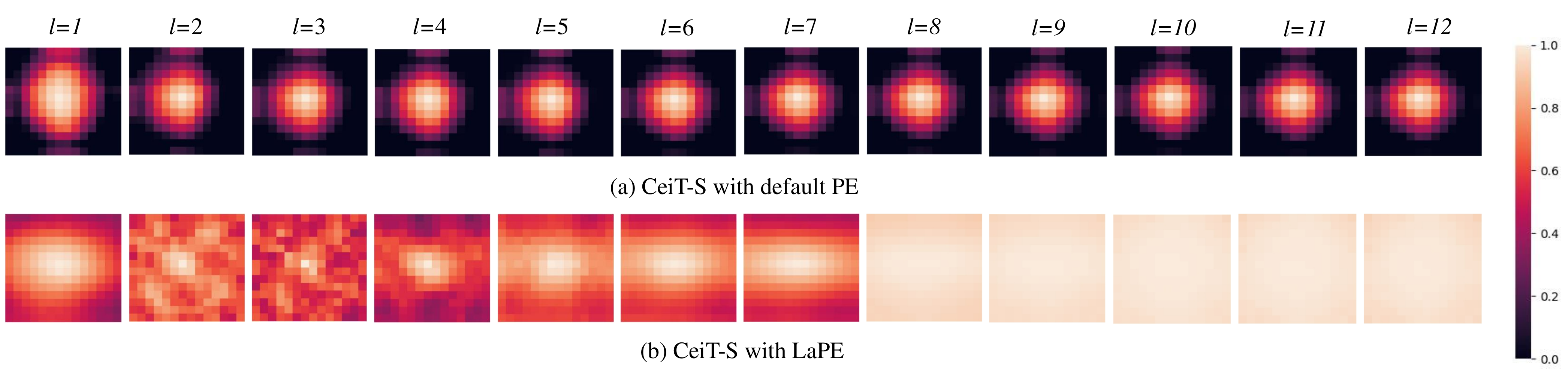}
   \caption{\textbf{Visualization of the position correlations at different layers for CeiT-S.} (a) The default position correlations are monotonic. (b) The position correlations adaptively change for each layer.}
   \label{fig:3}
\end{figure}

\section{Analyzing LaPE on More VTs By Visualization}
In the main body of this article, we provide selected layer's visualization of position information of T2T-ViT\cite{yuan2021tokens} and DeiT\cite{touvron2021training}. Here we supply the complete visualization for DeiT-Ti\cite{touvron2021training}, T2T-ViT-7\cite{yuan2021tokens} and CeiT-S\cite{yuan2021incorporating}.

% 我们训练了三个具有不同位置信息的T2T-ViT-7，其中一个是使用了原本的PE，另一个是对T2T的每一层都加入了LaPE，最后一个是只对前三层加入LaPE。我们对这三个模型中的位置信进行了可视化，结果如Fig1所示。通过观察，我们发现T2T with default PE第一层的位置信息被调整为2-D，而自第二层起，其位置信息仍为1-D且具有单调性，即每层的位置信息几乎不变。这是因为PE和token embedding共享了同一个LN，而该LN的仿射变换系数需要衡量这两种信息。而在T2T中，第一层LN的仿射变换学习去调整位置信息，而之后层LN的仿射变换则没有关注到位置信息上。
% 通过观察Fig1(b)我们可以发现，T2T-ViT-7的后几层将位置信息调节到
We train three T2T-ViT-7s\cite{yuan2021tokens} with different position information, including default PE, LaPE for each layer, and LaPE for the first 3 layers. As shown in Fig.~\ref{fig:1}, we visualize the position correlations of each layer. For T2T-ViT-7 with default PE (Fig.~\ref{fig:1}~(a)), we can see that the position correlation of the first layer is adjusted into 2-D correlation, while position correlations of the latter layers remain 1-D and keep monotonic, that is, the position correlations are nearly unchanged in the latter layers. This is because the PE and token embedding share the same layer normalization (LN), and the affine transformation coefficients of LN need to trade off between these two kinds of information. For T2T-ViT-7, the affine transformation coefficients of the first layer's LN learn to adjust the position information, while the coefficients of latter layers' LN do not pay attention to it. For T2T-ViT-7 with LaPE for each layer (Fig.~\ref{fig:1}~(b)), we can see that the position correlations are adjusted into 2-D and are hierarchical among layers. Meanwhile, we can see that the figures of the last few layers turn nearly all white, which means tokens are globally correlated. Therefore, we remove the LaPE for the last 4 layers, as it provides little position information for these layers, and the position correlations are shown in Fig.~\ref{fig:1}~(c). 

We also visualize the position correlations of DeiT-Ti\cite{touvron2021training} with and without LaPE. As shown in Fig.~\ref{fig:2}, DeiT-Ti with default PE shows monotonic position correlations, while DeiT-Ti with LaPE shows hierarchical position correlations. 

As shown in Fig.~\ref{fig:3}, we visualize the position correlations of CeiT-S\cite{yuan2021incorporating}. The CeiT-S with default PE shows monotonic position correlations same as previous models. However, the CeiT-S with LaPE shows slightly different position correlations from previous models. The LaPE-based position correlations are not exactly 2-D and do not completely follow the order from local to global. This is because CeiT uses Locally-Enhanced Feed-Forward Network (LeFF) to replace the MLP, and LeFF introduces locality information (containing position information) to models. Therefore, the Multi-Head Self-Attention (MSA) is not the only module containing the position information. Thus, the position correlations of CeiT-S with LaPE are adjusted into the shape in Fig.~\ref{fig:3}~(b).

% 对于CeiT-S来说，这个规律变得有些不同，即并非完全具有2-D特性，也并非完全遵循从local到global的秩序。这是由于CeiT使用了Locally-Enhanced Feed-Forward Network （LeFF）代替原有的MLP，而LeFF引入了locality information，因此MSA部分并非唯一引入位置信息的模块，所以LaPE中的position correlation学成了图3(b)中的形态。

\section{Training Settings for Image Classification Experiments}
ViT\_Lite, CVT, and CCT\cite{hassani2021escaping} on tiny datasets use the SGD\cite{rumelhart1986learning} as the optimizer, while DeiT\cite{touvron2021training}, T2T-ViT\cite{yuan2021tokens}, Swin\cite{liu2021swin} and CeiT\cite{yuan2021incorporating} on ImageNet-1K all use the Adamw\cite{loshchilov2017decoupled}. We list the hyper-parameters and settings used in our paper in Table\ref{tab:1}, which are the same as those in the original papers. 

\begin{table*}[t]
    \centering
    % \small
    \begin{tabular}{c|c|cccccc}
        \toprule[1.2pt]
        Dataset & Model & Learning Rate & \makecell[c]{Learning Rate\\Scheduler} & \makecell[c]{Weight\\Decay} &  \makecell[c]{Batch\\Size} & Epochs & \makecell[c]{Warm-up\\Epochs}\\
        \toprule[1.2pt]
        \multirow{8}{*}{ImageNet\cite{deng2009imagenet}} & \rule{0pt}{16pt} DeiT\cite{touvron2021training} & 5e-4 & \makecell[c]{cosine,\\min\_lr=1e-5} & 0.05 & 1024 & 300 & 5\\
        \cline{2-8}
        & \rule{0pt}{16pt} T2T-ViT\cite{yuan2021tokens} & 1e-3 & \makecell[c]{cosine,\\min\_lr=1e-5} & 0.03 & 1024 &
        \makecell[c]{300+10\\(cool\_down epochs)} & 10\\
        \cline{2-8}
        & \rule{0pt}{16pt} Swin\cite{liu2021swin} & 5e-4 & \makecell[c]{cosine,\\min\_lr=5e-6} & 0.05 & 512 & 300 & 20\\
        \cline{2-8}
        & \rule{0pt}{16pt} CeiT\cite{yuan2021incorporating} & 5e-4 & \makecell[c]{cosine,\\min\_lr=1e-5} & 0.05 & 1024 & 300 & 5\\
        \toprule[1.2pt]
        \multirow{6}{*}{Cifar-10\cite{krizhevsky2009learning}} & \rule{0pt}{16pt} ViT\_Lite\cite{hassani2021escaping} & 55e-5 & \makecell[c]{cosine,\\min\_lr=1e-5} & 0.06 & 128 & \makecell[c]{300+10\\(cool\_down epochs)} & 10\\
        \cline{2-8}
        & \rule{0pt}{16pt} CVT\cite{hassani2021escaping} & 55e-5 & \makecell[c]{cosine,\\min\_lr=1e-5} & 0.06 & 128 & \makecell[c]{300+10\\(cool\_down epochs)} & 10\\
        \cline{2-8}
        & \rule{0pt}{16pt} CCT\cite{hassani2021escaping} & 55e-5 & \makecell[c]{cosine,\\min\_lr=1e-5} & 0.06 & 128 & \makecell[c]{300+10\\(cool\_down epochs)} & 10\\
        \toprule[1.2pt]
        \multirow{6}{*}{Cifar-100\cite{krizhevsky2009learning}} & \rule{0pt}{16pt} ViT\_Lite\cite{hassani2021escaping} & 6e-4 & \makecell[c]{cosine,\\min\_lr=1e-5} & 0.06 & 128 & \makecell[c]{300+10\\(cool\_down epochs)} & 10\\
        \cline{2-8}
        & \rule{0pt}{16pt} CVT\cite{hassani2021escaping} & 6e-4 & \makecell[c]{cosine,\\min\_lr=1e-5} & 0.06 & 128 & \makecell[c]{300+10\\(cool\_down epochs)} & 10\\
        \cline{2-8}
        & \rule{0pt}{16pt} CCT\cite{hassani2021escaping} & 6e-4 & \makecell[c]{cosine,\\min\_lr=1e-5} & 0.06 & 128 & \makecell[c]{300+10\\(cool\_down epochs)} & 10\\
    	\bottomrule[1.2pt]
    \end{tabular}
\caption{\textbf{Experimental settings on ImageNet, CIFAR10 and CIFAR100.}}
\label{tab:1}
\end{table*}

\end{document}